\documentclass{article}

\PassOptionsToPackage{numbers,sort&compress}{natbib}
\usepackage[preprint]{neurips_2026}

\usepackage[utf8]{inputenc} 
\usepackage[T1]{fontenc}    
\usepackage{hyperref}       
\usepackage{url}            
\usepackage{booktabs}       
\usepackage{amsfonts}       
\usepackage{nicefrac}       
\usepackage{microtype}      
\usepackage{xcolor}         
\usepackage{amsmath}
\usepackage{booktabs}
\usepackage{multirow}
\usepackage{mathtools}
\usepackage{graphicx}
\usepackage{tabularx}
\usepackage{array}
\usepackage{amssymb}
\usepackage{pifont}
\usepackage{caption}
\usepackage{wrapfig}
\usepackage{enumitem}
\usepackage{algorithm}
\usepackage{algpseudocode}
\usepackage{subcaption}
\usepackage[most]{tcolorbox}
\usepackage{xcolor}
\newcommand{\cmark}{\ding{51}}
\newcommand{\xmark}{\ding{55}}

\definecolor{successgreen}{RGB}{0,128,0}
\definecolor{failred}{RGB}{180,0,0}
\definecolor{omitgray}{RGB}{120,120,120}
\newcolumntype{L}[1]{>{\raggedright\arraybackslash}p{#1}}
\newcolumntype{C}[1]{>{\centering\arraybackslash}p{#1}}

\usepackage[most]{tcolorbox}
\usepackage{xcolor}
\usepackage{enumitem}
\usepackage{tabularx}

\usepackage[most]{tcolorbox}
\usepackage{xcolor}

\definecolor{promptblue}{RGB}{0,0,140}
\definecolor{promptbg}{RGB}{246,247,255}
\definecolor{promptorange}{RGB}{220,120,35}

\newcommand{\strong}[1]{\textbf{\textcolor{promptblue}{#1}}}

\newcommand{\tagname}[1]{\textbf{\texttt{#1}}}
\newcommand{\pvar}[1]{\textcolor{promptorange}{\texttt{\symbol{123}\detokenize{#1}\symbol{125}}}}

\newtcolorbox{promptbox}[1]{
    enhanced,
    colback=promptbg,
    colframe=promptblue,
    colbacktitle=promptblue,
    coltitle=white,
    fonttitle=\bfseries,
    title={#1},
    arc=3mm,
    outer arc=3mm,
    boxrule=0.9pt,
    left=4mm,
    right=4mm,
    top=2mm,
    bottom=2mm,
    toptitle=1mm,
    bottomtitle=1mm,
    titlerule=0pt,
    before upper={
        \setlength{\parindent}{0pt}
        \setlength{\parskip}{0pt}
        \raggedright
    }
}

\usepackage[most]{tcolorbox}
\usepackage{xcolor}

\definecolor{memorybrown}{RGB}{165,82,0}
\definecolor{memorybg}{RGB}{255,250,244}
\definecolor{memoryorange}{RGB}{210,105,30}

\newtcolorbox{memorybox}[1]{
    enhanced,
    colback=memorybg,
    colframe=memorybrown,
    colbacktitle=memorybrown,
    coltitle=white,
    fonttitle=\bfseries,
    title={#1},
    arc=3mm,
    outer arc=3mm,
    boxrule=0.9pt,
    left=4mm,
    right=4mm,
    top=2mm,
    bottom=2mm,
    toptitle=1mm,
    bottomtitle=1mm,
    titlerule=0pt,
    before upper={
        \setlength{\parindent}{0pt}
        \setlength{\parskip}{0pt}
        \raggedright
    }
}

\title{HIPIF: Hierarchical Planning and Information Folding for Long-Horizon LLM Agent Learning}

\author{
    \textbf{Juncheng Diao\textsuperscript{1,2}$^{\ast}$,
    Zhicong Lu\textsuperscript{2}$^{\ast}$$^\dagger$,
    Peiguang Li\textsuperscript{1},
    Yongwei Zhou\textsuperscript{1}} \\
    \textbf{Changyuan Tian\textsuperscript{2},
    Qingbin Li\textsuperscript{2},
    Rongxiang Weng\textsuperscript{1},
    Jingang Wang\textsuperscript{1},
    Xunliang Cai\textsuperscript{1}} \\[2mm]
    \textsuperscript{1}Meituan \qquad
    \textsuperscript{2}University of Chinese Academy of Sciences \\[1mm]
    \texttt{diaojuncheng24@mails.ucas.ac.cn} \qquad
    \texttt{luzhicong21@mails.ucas.ac.cn}
}

\begin{document}

\footnotetext[1]{Equal contribution.}
\footnotetext[2]{Corresponding author.}

\maketitle

\begin{abstract}
While Large Language Models (LLMs) have demonstrated strong capabilities as autonomous agents across a wide range of tasks, their performance often degrades in multi-turn long-horizon agentic tasks. Existing methods have made progress through fine-grained credit assignment to alleviate long-horizon sparse rewards and hierarchical reinforcement learning to decompose tasks and reduce long-term dependency. 
However, these methods still do not directly address long-context interference, in which continuously growing histories weaken the agent's ability to track the global task state and impair subsequent reasoning and decision-making. Inspired by the way humans handle complex tasks through subgoal decomposition and completed progress summarization, we propose \textbf{Hi}erarchical \textbf{P}lanning and \textbf{I}nformation \textbf{F}olding (HIPIF) for long-horizon LLM agent learning. HIPIF trains the agent end-to-end to organize long-horizon execution around explicit subgoals while folding completed subgoal histories to reduce long-context interference. Furthermore, to stabilize subgoal-based planning and execution, HIPIF combines hierarchical reflection and subgoal-oriented process rewards to guide subgoal generation, transition, and execution, without relying on costly auxiliary models or task-specific expert trajectories. Extensive experiments on three publicly available agentic benchmarks demonstrate the validity of our method.
\end{abstract}

\section{Introduction}

Large Language Models (LLMs) have been a promising foundation for long-horizon agentic decision-making tasks benefiting from their ever-growing reasoning and planning abilities, where an agent must accomplish a high-level goal through multi-turn interaction with the environment~\citep{wangvoyager,yao2023react,shridharalfworld,wang2022scienceworld,puig2018virtualhome}. Despite this potential, compared with their success in single-step tasks, existing LLM agents remain far from satisfactory in complex long-horizon interactions. As noted by STEP-HRL~\citep{zhen2026hierarchical}, a key limitation of existing LLM agents~\citep{chen2021decision,ni2023transformers} is their reliance on an ever-growing observation-action history for each decision. In long-horizon interactions, the continuously growing context accumulates redundant information that weakens the agent's ability to track the global task state and impairs subsequent reasoning and decision-making~\citep{zhoumem1}.

Existing methods have made preliminary attempts to address this challenge. Prompt-based methods~\citep{yao2023react,yao2023tree,shinn2023reflexion,lin2023swiftsage,hu2025hiagent} and behavior cloning methods~\citep{chen2024agent,li2022pre} mainly rely on prompt engineering or expert trajectories to elicit reasoning, planning, reflection or context-folding abilities in LLM agents. However, they are not optimized through environmental feedback, which limits their adaptability across diverse environments and long-horizon interactions.
In contrast, reinforcement learning (RL) methods improve long-horizon agents from an optimization perspective by using environmental feedback to provide more fine-grained and reliable reward signals. For example, credit assignment methods~\citep{wang2025spa,wang2025steca,guosegment,fenggroup,lu2026hisr} alleviate sparse-reward challenges in long-horizon tasks through more precise step-level supervision, while hierarchical RL methods~\citep{hu2025divide,zhao2024epo,zhen2026hierarchical,peng2026hiper} reduce long-term dependency through task decomposition. 
Nevertheless, many existing RL methods rely on additional models for task decomposition or process-reward annotation, increasing pipeline complexity and limiting scalability across environments. 
More importantly, these methods rarely train the model to organize and fold ever-growing contexts and therefore cannot fundamentally resolve the state-tracking failure and reasoning degradation caused by long-context interference.

Inspired by the way humans handle long-horizon tasks through subgoal decomposition and completed progress summarization, we propose \textbf{Hi}erarchical \textbf{P}lanning and \textbf{I}nformation \textbf{F}olding (HIPIF) for long-horizon LLM agent learning. HIPIF trains the agent end-to-end to organize long-horizon execution around explicit subgoals and fold the execution histories of completed subgoals, thereby reducing long-context interference. To stabilize subgoal-based planning and execution, HIPIF introduces hierarchical reflection to improve subgoal transition judgment and guide either subgoal proposal or current subgoal execution. Furthermore, to alleviate sparse rewards in long-horizon subgoal-based training, HIPIF introduces subgoal-oriented process rewards to correct inappropriate subgoals and ineffective execution behaviors within subgoals.

Extensive experimental results on three publicly available agentic benchmarks and case studies demonstrate the effectiveness of HIPIF. 
Further efficiency analyses show that HIPIF achieves lower token usage in long-horizon interactions while avoiding task-specific expert trajectories and additional auxiliary models. 
In summary, our main contributions are as follows.
\begin{itemize}
    \item We propose \textbf{Hi}erarchical \textbf{P}lanning and \textbf{I}nformation \textbf{F}olding (HIPIF) for long-horizon LLM agent learning, which trains the model to organize long-horizon execution around explicit subgoals and fold the histories of completed subgoals to reduce long-context interference.
    \item To stabilize subgoal-based planning and execution, we introduce hierarchical reflection and subgoal-oriented process rewards to improve subgoal completion judgment, subgoal content assessment, and subgoal execution correction.
    \item Extensive experimental results on three publicly available agentic benchmarks, efficiency analyses, and case studies demonstrate the effectiveness and efficiency of HIPIF. 
\end{itemize}

\section{Related Work}
\label{gen_inst}

\paragraph{LLM Agents.}
Large Language Models (LLMs) have been widely used as agents in interactive decision-making tasks~\citep{vaswani2017attention,wangvoyager,yao2023react}. Early studies primarily adopt prompt-based formulations, where agents externalize intermediate decision processes to support multi-step decision-making, such as Chain-of-Thought~\citep{wei2022chain}, ReAct~\citep{yao2023react} and Reflexion~\citep{shinn2023reflexion}.
To improve LLM agents in long-horizon tasks, several methods introduce memory mechanisms~\citep{zhang2025survey,xu2026amem,sarch2024vlm}. For example, HiAgent~\citep{hu2025hiagent} uses prompts to guide subgoal decomposition and history folding. However, these mechanisms are usually based on hand-crafted prompts or system designs without environmental feedback, thus unreliable in complex long-horizon tasks.
Another line of work learns agent policies from expert trajectories through behavior cloning or supervised fine-tuning~\citep{chen2024agent,li2022pre}. However, such methods heavily depend on task-specific expert trajectories, which are costly and face limited scalability across environments.

\paragraph{Reinforcement Learning in LLM Agents.}
Reinforcement learning (RL) provides a mechanism for optimizing LLM agents through environmental interaction and reward feedback~\citep{xu2024language,wang2025ragen,ni2023transformers}. Existing work applies PPO~\citep{schulman2017proximal}, GRPO~\citep{shao2024deepseekmath}, RLOO~\citep{ahmadian2024back}, or preference-based optimization~\citep{rafailov2023direct} to LLM agents, enabling the model to improve its behavior from environment signals. 
There are also studies that focus on fine-grained reward assignment for long-horizon agent training~\citep{wang2025spa,wang2025steca,guosegment}, since final task rewards are often sparse and delayed. These methods provide more localized training signals through turn-level process reward models or step-level advantage estimation, with representative examples including GiGPO~\citep{fenggroup} and HiSR~\citep{lu2025piper}. However, they mainly improve credit assignment within trajectories while still making decisions based on the full observation-action history. As a result, these methods still lack explicit task-stage organization and context management, and therefore cannot fundamentally mitigate the reasoning degradation caused by long contexts.
Recent work also trains memory or context compression mechanisms with RL~\citep{li2026deepagent,sarch2024vlm}, such as FoldGRPO~\citep{sun2025scaling}, A-Mem~\citep{xu2026amem}, and AgentFold~\citep{ye2026agentfold}. These methods recognize that memory writing, retrieval, or context compression can be optimized through reinforcement learning. Nevertheless, they mainly focus on compressing long contexts rather than systematically improving decision reliability in long-horizon agents.
In addition, hierarchical RL methods introduce hierarchical structures for long-horizon tasks by decomposing complex goals into subgoals and optimizing policies accordingly~\citep{hu2025divide,zhao2024epo,zhouarcher}. For example, HiPER~\citep{peng2026hiper} focuses on subgoal proposal and subgoal-level credit assignment, while STEP-HRL~\citep{zhen2026hierarchical} improves long-horizon agent training from the perspectives of subgoal modeling and context compression. These methods demonstrate the value of subgoal for complex interactive tasks. However, many existing methods still rely on auxiliary models  or task-specific expert trajectories for subgoal generation, context compression, or critic estimation, which increases training pipeline complexity and limits scalability across environments.

\section{Methodology}
\label{headings}

\begin{figure}[htbp]
  \centering
  \includegraphics[width=\textwidth]{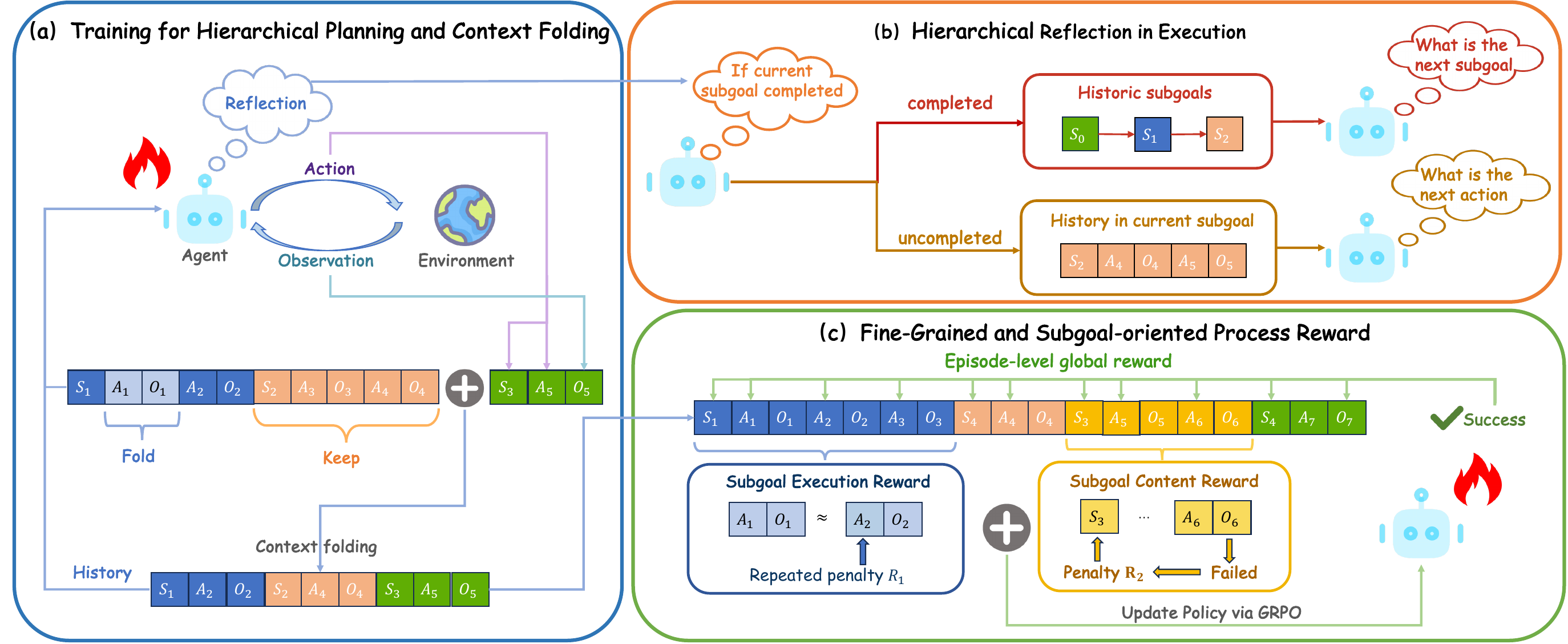}
  
  \caption{Overview of the design of HIPIF. (a): End-to-End Training for Hierarchical Planning and Information Folding. (b): Hierarchical reflection. (c): Subgoal-oriented process rewards.}
  \label{fig:method}
\end{figure}

In this section, we present the overall design of HIPIF. 
As illustrated in Figure~\ref{fig:method}(a), HIPIF adopts end-to-end training for hierarchical planning and information folding. 
To stabilize subgoal-based planning and execution, Figure~\ref{fig:method}(b) introduces a hierarchical reflection mechanism. 
Finally, Figure~\ref{fig:method}(c) shows the subgoal-oriented process rewards for both subgoal generation and execution within the subgoal. The complete training pipeline is summarized in Algorithm~\ref{alg:hipif}.

\subsection{End-to-End Training for Hierarchical Planning and Information Folding}
\label{sec:grpo_subgoal_folding}
To reduce long-context interference while making subgoal-based execution trainable, we introduce Subgoal-Level Information Folding  and GRPO Training for Subgoal-Centric Decisions.

\textbf{Hierarchical Planning and Information Folding.}
In conventional multi-turn agent tasks, an LLM agent typically follows a history-conditioned formulation. At interaction step $t$, the policy conditions on the full accumulated trajectory are:
\begin{equation}
\tau_t=(c,o_1,a_1,o_2,a_2,\ldots,o_t),
\label{eq:full_history}
\end{equation}
where $c$ denotes the task, $o_t$ is the observation returned by the environment, and $a_t$ is the model response, usually consisting of thought-action pairs, which adheres to ReAct~\citep{yao2023react}.
In long-horizon tasks, full interaction histories continuously accumulate redundant information. Such context noise weakens the agent's awareness of the current task stage and degrades its decision-making ability.

We draw inspiration from the way humans handle long-horizon tasks, in which they decompose complex objectives into subgoals and summarize completed progress. For instance, in PICK2 tasks from ALFWorld, once the subgoal of moving the first object has been completed, the agent should fold the corresponding execution history of the first object and focus on moving the second object. Retaining the full execution history of the first object may instead introduce context interference and confuse the model's subsequent decisions.

Motivated by this observation, HIPIF organizes long-horizon interaction around explicit subgoals and folds the execution histories of completed subgoals. The model first proposes an initial subgoal according to the task description. Given the current subgoal, the model then repeatedly generates actions to execute it, while the environment returns a new observation after each action. 
Once the model judges that the current subgoal has been completed or should be terminated, HIPIF folds the execution history of this subgoal and proposes the next subgoal.
At each decision step, HIPIF maintains a compact working context by combining folded global progress with detailed local execution history. Formally, at step $j$ of the current subgoal $g_k$, the context provided to the policy is:
\begin{equation}
C_{k,j} = [c;\mathcal{H}_{<k};g_k;\mathcal{T}_{k,j}],
\label{eq:working_context}
\end{equation}
where $c$ is the task description, $\mathcal{H}_{<k}$ denotes the folded records of completed subgoals before $g_k$, and $\mathcal{T}_{k,j}$ denotes the action-observation history within $g_k$ up to step $j$.
Detailed implementations and examples are provided in Appendix~\ref{app:prompt_memory}.

\textbf{End-to-End Training for Subgoal-Centric Decisions.}
Using prompts alone for subgoal decomposition and history folding is insufficient, as the model lacks feedback-driven training on what subgoals to propose and how to execute them reliably. Therefore, HIPIF treats subgoal-centric decision-making as a learnable policy behavior and optimizes it through environmental feedback. 
Given the folded working context $C_{k,j}$ and the reflection $\xi_{k,j}$ from Section~\ref{sec:hierarchical_reflection}, the model generates the next decision:
\begin{equation}
y_{k,j}
\sim
\pi_\theta(\cdot \mid C_{k,j}, \xi_{k,j}),
\label{eq:subgoal_policy}
\end{equation}

Specifically, when the model decides to continue the current subgoal, the next action $a_{k,j}$ is extracted from the \texttt{action} field and executed under $g_k$; when the model decides to terminate the current subgoal, the new subgoal $g_{k+1}$ and its first action $a_{k+1,1}$ are extracted from the \texttt{subgoal} and \texttt{action} fields, respectively. Once the transition happens, the completed subgoal $g_k$ is folded into a compact record $[ g_k,o_k^{\mathrm{end}}]$ and appended to the folded history.

To train these behaviors, HIPIF adopts GRPO with the subgoal-oriented process rewards defined in Section~\ref{sec:process_rewards}. 
For each task instruction, we sample a group of $M$ trajectories from the old policy, recording the context at each step together with the generated reflections, actions, and possible subgoals. 
Following verl-agent~\citep{fenggroup}, instead of assigning a single trajectory-level advantage to all decisions, we compute step-level returns by combining the final task outcome with the subgoal-oriented process rewards, and then normalize these returns within the sampled group following the procedure in Section~\ref{sec:process_rewards}, which yields a step-level advantage $\hat{A}_{t}^{(m)}$ for each decision step.
The policy is then optimized with the clipped GRPO objective:
\begin{equation}
\resizebox{0.92\linewidth}{!}{$
\mathcal{L}_{\mathrm{GRPO}}(\theta)
=
-\mathbb{E}_{m,t}
\left[
\min
\left(
\frac{
\pi_\theta(\zeta_t^{(m)} \mid C_t^{(m)})
}{
\pi_{\theta_{\mathrm{old}}}(\zeta_t^{(m)} \mid C_t^{(m)})
}
\hat{A}_{t}^{(m)},
\;
\mathrm{clip}
\left(
\frac{
\pi_\theta(\zeta_t^{(m)} \mid C_t^{(m)})
}{
\pi_{\theta_{\mathrm{old}}}(\zeta_t^{(m)} \mid C_t^{(m)})
},
1-\epsilon_{\mathrm{clip}},
1+\epsilon_{\mathrm{clip}}
\right)
\hat{A}_{t}^{(m)}
\right)
\right]
$}
\label{eq:grpo_objective}
\end{equation}
where $\pi_\theta$ and $\pi_{\theta_{\mathrm{old}}}$ denote the current and old policies, respectively. 
$\zeta_t^{(m)}$ denotes the complete structured decision sequence generated autoregressively at decision step $t$, including the rationale, completion judgment, branch-specific reflection, and the final action or subgoal decision. All generated response tokens in $\zeta_t^{(m)}$ are optimized jointly under the GRPO objective.

\subsection{Hierarchical Reflection in Execution}
\label{sec:hierarchical_reflection}

Although the previous section formulates the subgoal proposal and context folding as part of the RL training process, it also introduces new challenges: in the early stage of agent training, it is often challenging for the model to simultaneously propose meaningful subgoals, determine when to switch subgoals, and execute the current subgoal. While external models or annotated trajectories can provide additional guidance, they incur high costs and limit the scalability on different environments.

To address this issue, we propose a hierarchical reflection mechanism during rollout, motivated by both temporal abstraction in hierarchical decision-making~\citep{sutton1999between} and Reflexion~\citep{shinn2023reflexion}. 
Reflection enables the agent to explicitly assess task progress and diagnose failure patterns before making the next decision. By embedding this reflective process into a temporally abstracted subgoal structure, HIPIF further turns reflection into a control mechanism for subgoal termination, transition, and execution.

Specifically, following temporal abstraction in hierarchical decision-making~\citep{sutton1999between}, we treat each subgoal $g_k$ as a stage-level control unit. Rather than corresponding to a single atomic action, a subgoal specifies the agent's local objective and constrains action generation over multiple interaction steps. After each execution step, the model first assesses whether the current subgoal has been completed. Let $\mathcal{H}_{<k}$ denote the folded history before subgoal $g_k$, $h_{k,t}$ denote the action-observation history within the current subgoal up to step $t$, and $o_t$ denote the current observation. The reflection module then generates a completion judgment together with its reasoning process:
\begin{equation}
(\eta_{k,t}, z_{k,t})
\sim
\mathcal \pi_{\theta}
\left(
\cdot \mid \mathcal{H}_{<k}, g_k, h_{k,t}, o_t
\right),
\end{equation}
where $z_{k,t}\in\{0,1\}$ indicates whether the current subgoal is completed, and $\eta_{k,t}$ denotes the reasoning process generated by the model. Following the intuition of chain-of-thought reasoning~\citep{wei2022chain}, the reasoning process encourages the model to examine the evidence behind its completion judgment and  improves the accuracy of subgoal-state assessment compared with directly predicting a binary label.

Based on this completion judgment, HIPIF branches into different generation modes:
\begin{equation}
\xi_{k,t} \sim
\begin{cases}
\pi_\theta\left(
\cdot \mid \mathcal{G}_{\leq k}, \eta_{k,t}
\right),
& z_{k,t}=1, \\[2mm]
\pi_\theta\left(
\cdot \mid g_k, h_{k,t}, o_t, \eta_{k,t}
\right),
& z_{k,t}=0.
\end{cases}
\label{eq:branch_generation}
\end{equation}
where $\xi_{k,t}$ denotes the agent's reflection for next output, which can be either about the next subgoal $g_{k+1}$ or about the next action $a_t$ depending on the completion judgment $z_{k,t}$.
Here, $\mathcal{G}_{\leq k}=\{g_1,\ldots,g_k\}$ is the sequence of previously folded subgoals.
When $z_{k,t}=1$, the model reflects on the previous folded subgoal histories $\mathcal{G}_{\leq k}$ to propose the next subgoal $g_{k+1}$, helping the model avoid redundant subgoals and identify the current state. When $z_{k,t}=0$, the model reflects based on the current subgoal $g_k$ and its execution history $h_{k,t}$ to identify the effective action, encouraging the model to avoid the invalid attempts. These reflections serve as the reasoning basis for Section~\ref{sec:grpo_subgoal_folding}. 
Detailed implementations are provided in Appendix~\ref{app:prompt_memory}.

\subsection{Fine-Grained and Subgoal-oriented Process Rewards}
\label{sec:process_rewards}

RL training for subgoal proposal and information folding faces sparse-reward challenges. A failed rollout may result from an inappropriate subgoal or ineffective execution under the current subgoal, while the final task reward alone struggles to distinguish these errors. Therefore, we design subgoal-oriented process rewards to provide more localized supervision for both subgoal content and subgoal execution. To avoid extra computational overhead and reduce the risk of reward hacking, we adopt rule-based process rewards that penalize only steps clearly identified as erroneous from environment feedback. Detailed implementations and analysis are provided in Appendix~\ref{app:training_details} and ~\ref{app:sensitivity_analysis}.

\textbf{Subgoal Content Reward.}
The first type of process reward evaluates subgoal content. 
Since the model is trained without expert subgoal annotations, it may generate subgoals that are not grounded in the current environment, making them difficult to execute and potentially misleading subsequent action generation. Therefore, we propose $r_t^{\mathrm{gr}}$ to penalize subgoals that refers to objects or receptacles absent from the available environment context. In addition, we identify unreliable subgoals in successful trajectories. If a trajectory eventually succeeds, it should contain a sequence of subgoals that can support task completion. Therefore, for successful trajectories, we further apply $r_t^{\mathrm{term}}$ to penalize subgoals whose terminal observation indicates execution failure, such as ``Nothing happens'', thereby exposing erroneous subgoals that may otherwise be masked by eventual task success.

\paragraph{Subgoal Execution Reward.}
The second type of process reward targets execution errors within the execution of a subgoal. Even when a subgoal is reasonable, the agent may still fail to execute it effectively. 
A common failure pattern is a loop within subgoal execution, where the agent repeatedly produces the same action and receives the same observation under the current subgoal. 
We therefore define the execution penalty $r_t^{\mathrm{exec}}$ on such repeated action-observation pairs under the same subgoal, which indicates that the agent is not making progress. In addition, we introduce a format penalty $r_t^{\mathrm{fmt}}$ to ensure valid structured outputs, penalizing the model when it omits or mismatches the required tag.

\paragraph{Process Reward Assignment.}
After defining the subgoal-content and execution-related penalties, we assign process feedback at the step level. 
The total process reward at step $t$ is defined as
\begin{equation}
r_t^{\mathrm{proc}}
=
r_t^{\mathrm{gr}}
+
r_t^{\mathrm{term}}
+
r_t^{\mathrm{exec}}
+
r_t^{\mathrm{fmt}},
\end{equation}

We then combine this local process feedback with the final task outcome to construct a step-level training score. 
Given the terminal environment reward $R_{\mathrm{env}}$, the score for step $t$ is defined as
\begin{equation}
S_t
=
R_{\mathrm{env}}
+
r_t^{\mathrm{proc}},
\end{equation}
where $S_t$ serves as a lightweight scoring signal that broadcasts the trajectory-level success outcome to each decision step while applying local penalties to clearly erroneous subgoal or execution behaviors.

For policy optimization, we follow the group-relative normalization used in GIGPO~\citep{fenggroup}. 
For the same instruction, we sample a group of trajectories and compute the step-level scores for all stored decision steps. 
The normalized step-level advantage is computed as
\begin{equation}
\hat{A}_t
=
\frac{S_t-\mu_S}{\sigma_S+\epsilon},
\end{equation}
where $\mu_S$ and $\sigma_S$ are the group mean and standard deviation of step-level scores.

\section{Experiments}
\label{exp}

\subsection{Experimental Settings}

\textbf{Benchmarks.} To systematically evaluate the effectiveness of the proposed method, we conduct experiments on three publicly available interactive agent benchmarks, including ALFWorld \citep{shridharalfworld}, VirtualHome \citep{puig2018virtualhome} and ScienceWorld \citep{wang2022scienceworld}. Specifically, for ALFWorld, following prior work \citep{wang2025spa,wang2025steca,fenggroup}, we adopt the dataset version constructed by \citet{song2024trial}. For VirtualHome, we further correct clearly erroneous examples based on the version provided by \citet{wang2025spa,wang2025steca}. For ScienceWorld, we use the same experimental setting as previous work~\citep{zhen2026hierarchical}. Across all benchmarks, at each interaction step, the agent receives an observation from the environment and generates the next action accordingly. The interaction continues until the task is successfully completed or a predefined maximum number of steps is reached, after which the environment returns the final task outcome. Additional details on the benchmarks are provided in Appendix~\ref{app:datasets}.

\textbf{Baselines.}
For the three benchmarks, we compare our approach with a range of competitive baselines: 
(1) Closed-source LLMs: GPT-4o~\citep{hurst2024gpt} and Gemini-2.5-Pro~\citep{team2023gemini}, which represent powerful capabilities in interactive reasoning and decision-making. 
(2) Prompting agents: ReAct~\citep{yao2023react}, Reflexion~\citep{shinn2023reflexion}, and HiAgent~\citep{hu2025hiagent} which rely on in-context prompting to guide test-time multi-turn behavior without parameter updates. 
(3) Credit-assignment-oriented RL training methods: PPO~\citep{schulman2017proximal} uses a learned value function for advantage estimation, while RLOO~\citep{ahmadian2024back} and GRPO~\citep{shao2024deepseekmath} estimate relative advantages from grouped samples without an additional critic; GiGPO~\citep{fenggroup} and RL-GCD~\citep{liu2025gradient} further introduce step-level advantage estimation for finer-grained credit assignment. 
(4) Hierarchical RL methods: Hiper~\citep{peng2026hiper} focus on subgoal-level credit assignment for multi-turn agentic RL. GLIDER~\citep{hu2025divide} and STEP-HRL~\citep{zhen2026hierarchical} combine subgoal modeling with supervised fine-tuning and offline RL. HiAgent+GRPO~\citep{hu2025hiagent,shao2024deepseekmath} is included to assess the effect of applying GRPO~\citep{shao2024deepseekmath} to a prompt-based subgoal framework.

\begin{figure*}[htbp]
\centering
\begin{subfigure}[b]{0.32\textwidth}
    \centering
    \includegraphics[width=\textwidth]{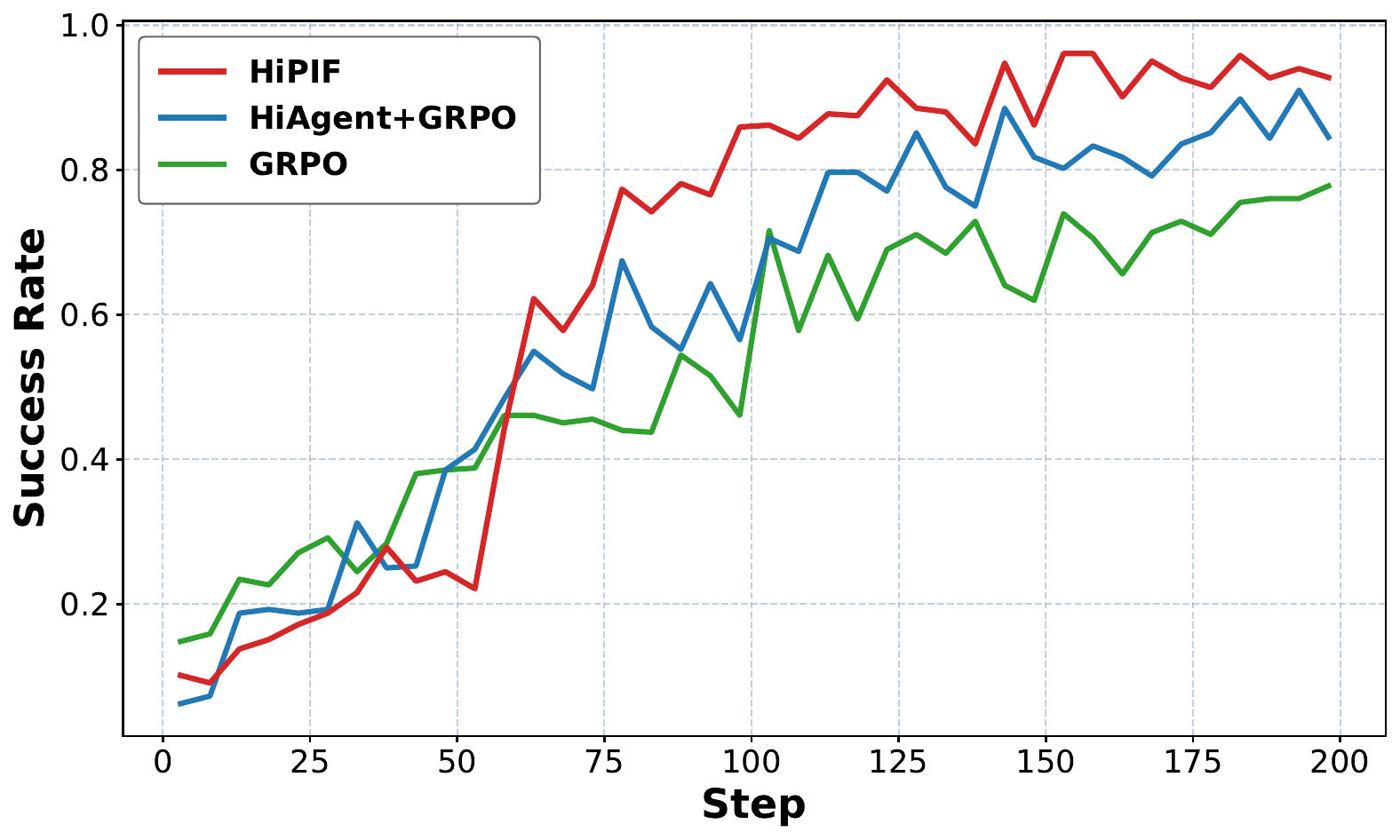}
    \caption{ALFWorld}
\end{subfigure}
\hfill
\begin{subfigure}[b]{0.32\textwidth}
    \centering
    \includegraphics[width=\textwidth]{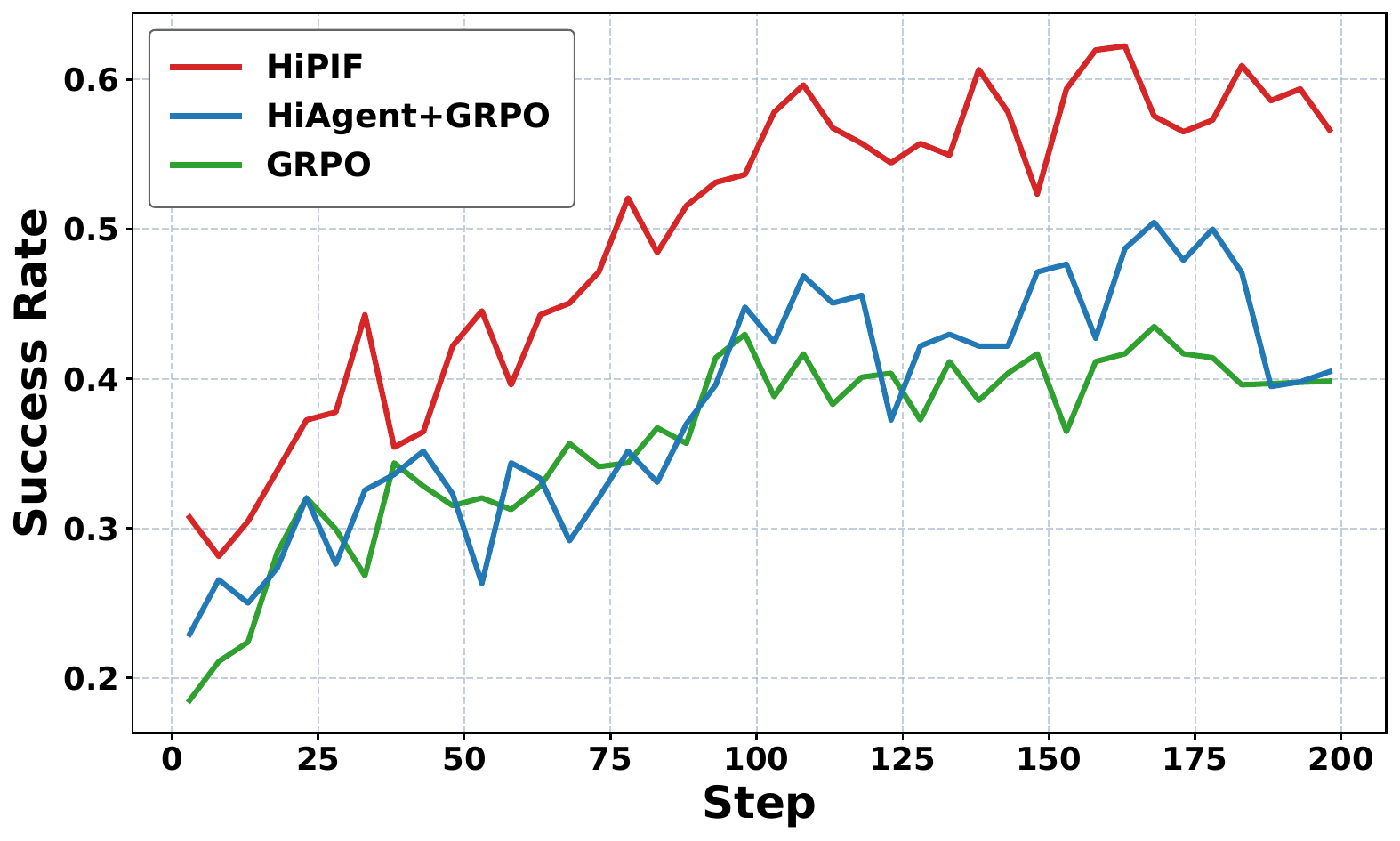}
    \caption{VirtualHome}
\end{subfigure}
\hfill
\begin{subfigure}[b]{0.32\textwidth}
    \centering
    \includegraphics[width=\textwidth]{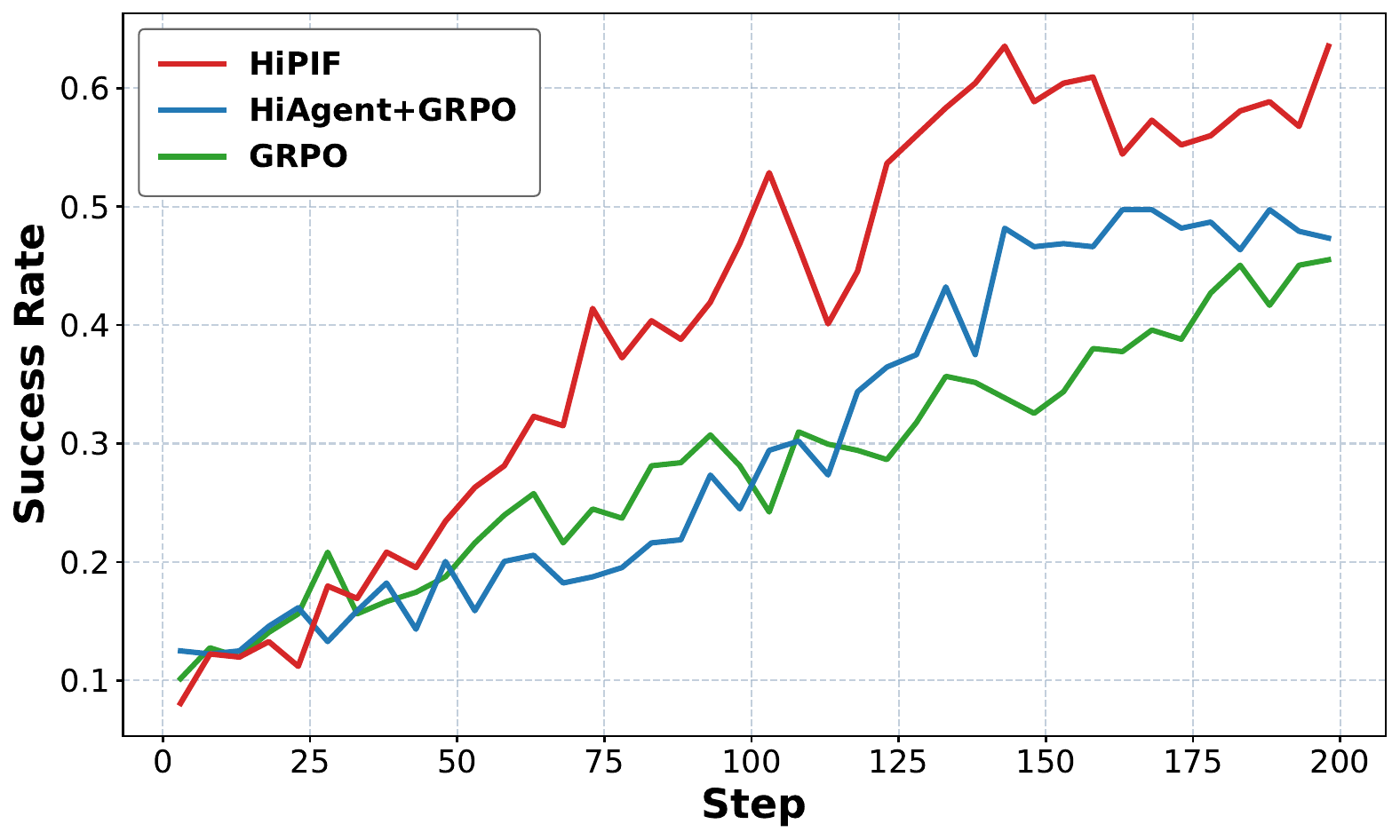}
    \caption{ScienceWorld}
\end{subfigure}
\caption{Validation success-rate curves of 3B models on three benchmarks.}
\label{fig:val_curve_3b}
\end{figure*}

\textbf{Training details.}
We use Qwen2.5-3B-Instruct and Qwen2.5-7B-Instruct~\citep{yang2024qwen2} as our base models. 
All RL experiments are implemented with the verl-agent framework~\citep{fenggroup}. 
For a fair comparison, all RL-based methods, including our method and the RL baselines, use the same hyperparameter configuration following GiGPO~\citep{fenggroup}. 
All experiments are conducted on 8 NVIDIA A100 80GB GPUs. 
Full training settings and hyperparameter details are provided in Appendix~\ref{app:training_details}.

\begin{table}[htbp]
\centering
\caption{Evaluation results on three benchmarks, where all reported values are success rates. \textit{Avg} denotes the average score on ALFWorld. PE and RL indicate methods of prompt engineering and reinforcement learning.Best and runner-up results are marked in \textbf{bold} and \underline{underline}, respectively.}
\label{tab:evaluation_results}
\vspace{1mm}
\scriptsize
\setlength{\tabcolsep}{2.0pt}
\renewcommand{\arraystretch}{0.96}
\resizebox{0.98\textwidth}{!}{%
\begin{tabular}{llccccccccc}
\toprule
\multirow{2}{*}{Methods} & \multirow{2}{*}{Type} & \multicolumn{7}{c}{Alfworld $\uparrow$} & \multirow{2}{*}{Virtualhome $\uparrow$} & \multirow{2}{*}{Scienceworld $\uparrow$} \\
\cmidrule(lr){3-9}
& & PICK $\uparrow$ & CLEAN $\uparrow$ & HEAT $\uparrow$ & COOL $\uparrow$ & LOOK $\uparrow$ & PICK2 $\uparrow$ & \textit{Avg} $\uparrow$ & & \\
\midrule
\multicolumn{11}{l}{\textbf{Closed-Source Model}} \\
\midrule
GPT4o & PE & 75.3 & 60.8 & 31.2 & 56.7 & 21.6 & 49.8 & 48.0 & 20.8 & 47.1 \\
Gemini2.5pro & PE & 92.8 & 63.3 & 32.4 & 34.5 & 26.6 & 58.7 & 60.3 & 31.7 & 54.7 \\
GPT4o (Hiagent) & PE & 79.4 & 42.6 & 18.9 & 26.2 & 27.6 & 30.8 & 42.7 & 21.1 & 32.0 \\
Gemini2.5pro (Hiagent) & PE & 92.5 & 56.0 & 22.2 & 38.9 & 28.6 & 38.2 & 47.3 & 28.1 & 42.8 \\
\midrule
\multicolumn{11}{l}{\textbf{Zero-Shot Prompting}} \\
\midrule
Qwen2.5-3B-Instruct & PE & 12.5 & 0.0 & 0.0 & 0.0 & 0.0 & 0.0 & 2.3 & 10.2 & 7.8 \\
React & PE & 65.1 & 12.5 & 8.3 & 10.0 & 8.3 & 0.0 & 17.2 & 14.1 & 10.2 \\
Reflexion & PE & 68.6 & 38.5 & 37.0 & 25.0 & 32.0 & 4.2 & 37.1 & 18.2 & 16.4 \\
Hiagent & PE & 52.5 & 2.3 & 3.5 & 4.7 & 18.8 & 0.0 & 13.6 & 18.8 & 10.8 \\
\midrule
\multicolumn{11}{l}{\textbf{Credit-assignment-oriented RL}} \\
\midrule
PPO & RL & 83.3 & 87.1 & 73.9 & 61.9 & 77.8 & 47.1 & 73.9 & 51.2 & 50.3 \\
RLOO & RL & 83.3 & 87.1 & 73.9 & 61.9 & 77.8 & 47.1 & 73.9 & 47.1 & 38.4 \\
GRPO & RL & 87.5 & 70.9 & 82.6 & 57.1 & 88.9 & 52.9 & 73.4 & 46.9 & 43.1 \\
RL-GCD & RL & \textbf{100.0} & 90.6 & 75.2 & 72.1 & 63.6 & 40.9 & 78.4 & 42.9 & 42.3 \\
GIGPO & RL & 95.0 & 96.4 & \textbf{100.0} & 88.5 & \textbf{100.0} & \underline{85.7} & \underline{93.8} & \underline{60.9} & 58.1 \\
\midrule
\multicolumn{11}{l}{\textbf{Hierarchical RL}} \\
\midrule
Hiper & RL & \underline{97.4} & 96.2 & 88.9 & 83.3 & 81.8 & 70.0 & 88.3 & 45.5 & 42.6 \\
GLIDER & RL & 83.3 & 87.1 & 78.3 & 71.4 & 77.8 & 41.2 & 75.4 & -- & 56.9 \\
STEP-HRL & RL & \textbf{100.0} & \textbf{100.0} & 90.0 & \underline{89.7} & \textbf{100.0} & 77.8 & 92.9 & -- & \underline{61.8} \\
Hiagent+GRPO & RL & 85.3 & \textbf{100.0} & \textbf{100.0} & 77.3 & 77.8 & 76.9 & 85.9 & 48.9 & 46.1 \\
\midrule
HIPIF & RL & \textbf{100.0} & \underline{96.6} & \underline{96.3} & \textbf{96.2} & \underline{93.2} & \textbf{95.2} & \textbf{96.1} & \textbf{63.3} & \textbf{64.8} \\
\bottomrule
\end{tabular}
}
\vspace{-4mm}
\end{table}

\subsection{ Experimental Results}

Table~\ref{tab:evaluation_results} reports the overall performance of different methods on three embodied agent benchmarks.  We can observe that: 
\textbf{(1) Limitations of advanced closed-source models.} 
Even advanced closed-source models such as GPT-4o and Gemini-2.5-Pro struggle with long-horizon tasks when using standard ReAct prompting and HiAgent prompting, revealing that model scale and general reasoning ability are insufficient to address context interference and goal forgetting in multi-step interactions. 
\textbf{(2) Zero-shot prompting remains insufficient.} ReAct and Reflexion achieve only 17.2 and 37.1 average scores on ALFWorld, respectively. HiAgent adopts a subgoal decomposition and context folding paradigm, but obtains an even lower score than ReAct and Reflexion. This indicates that base models cannot reliably perform hierarchical planning  and information folding without training.  
\textbf{(3) Bottlenecks of credit-assignment-oriented RL methods.}
Credit-assignment-oriented RL improves over prompting baselines, but remains limited by the lack of explicit subgoal structure and context management. GiGPO improves fine-grained credit assignment through step-level advantage comparison and performs best among credit-assignment-oriented RL baselines, yet it still underperforms HIPIF across all three benchmarks. The gap is especially clear on complex tasks such as PICK2 where HIPIF improves over GiGPO from 85.7 to 95.2. This suggests that credit assignment alone cannot address long-context-induced degradation in reasoning and decision-making, while its complementarity to our framework is further analyzed in Appendix~\ref{app:val_curve_gigpo}.
\textbf{(4) Challenges in hierarchical RL methods.}
Existing hierarchical RL methods explicitly introduce hierarchical structures, but still suffer from several limitations. HiPER models subgoals with subgoal-level credit assignment, but lacks effective context management and hierarchical reflection. STEP-HRL depends on expert trajectories and external models for subgoal generation and context compression, which substantially increases training and inference costs. Despite these additional resources, it still underperforms HIPIF on three benchmarks.  
Although HiAgent+GRPO strengthens subgoal-and-folding with RL, its lower performance than HIPIF across all benchmarks suggests that effective long-horizon execution also requires hierarchical reflection and fine-grained rewards for subgoal transitions and execution.
\textbf{(5) Overall advantages of HIPIF.}
Compared with all baselines, HIPIF achieves the best performance. 
Figure~\ref{fig:val_curve_3b} further reveals two key advantages of HIPIF. 
First, learning to organize long-horizon interaction at the subgoal level and compress completed execution of subgoals can substantially improve model performance. 
Second, subgoal-based training may underperform in the early stage because it introduces a more complex decision process. 
The growing advantage of HIPIF over the other two methods during training demonstrates that hierarchical reflection and fine-grained process-level rewards effectively stabilize subgoal-based RL training.
\begin{table}[htbp]
\centering
\caption{Ablation studies of our proposed method across different model scales on three embodied agent benchmarks. Best results within each model size group are marked in \textbf{bold}.}
\label{tab:multi_scale_ablation}
\vspace{2mm}
\setlength{\tabcolsep}{2.5pt}
\footnotesize
\begin{tabular}{lccccccccc}
\toprule
\multirow{2}{*}{\textbf{Methods}} & \multicolumn{7}{c}{\textbf{ALFWorld} $\uparrow$} & \multirow{2}{*}{\textbf{VirtualHome} $\uparrow$} & \multirow{2}{*}{\textbf{ScienceWorld} $\uparrow$} \\
\cmidrule(lr){2-8}
& PICK & CLEAN & HEAT & COOL & LOOK & PICK2 & \textit{Avg} & & \\
\midrule
\multicolumn{10}{c}{\textit{Base Model Scale: 3B}} \\
\midrule
Ours (3B) & \textbf{100.0} & \textbf{96.6} & \textbf{96.3} & \textbf{96.2} & 93.2 & \textbf{95.2} & \textbf{96.1} & \textbf{63.3} & \textbf{64.8} \\
\quad - w/o Reflection & 91.7 & 87.1 & 78.3 & 66.7 & \textbf{100.0} & 64.7 & 87.8 & 53.9 & 49.2 \\
\quad - w/o Reward & 100.0 & 96.4 & 94.4 & 75.0 & 83.2 & 92.6 & 90.3 & 57.9 & 57.0 \\
\quad - w/o Subgoal & 83.3 & 87.1 & 73.9 & 61.9 & 77.8 & 47.1 & 73.9 & 46.9 & 43.1 \\
\midrule
\multicolumn{10}{c}{\textit{Base Model Scale: 7B}} \\
\midrule
Ours (7B) & \textbf{100.0} & \textbf{100.0} & \textbf{100.0} & \textbf{100.0} & \textbf{100.0} & \textbf{95.0} & \textbf{99.2} & \textbf{68.8} & \textbf{71.5} \\
\quad - w/o Reflection & 92.3 & \textbf{100.0} & 94.7 & \textbf{100.0} & 94.1 & 78.9 & 93.8 & 61.2 & 66.4 \\
\quad - w/o Reward & \textbf{100.0} & 97.5 & 75.0 & \textbf{100.0} & 82.5 & 91.7 & 91.4 & 62.8 & 69.2 \\
\quad - w/o Subgoal & 88.9 & 88.5 & 88.0 & 85.7 & 94.1 & 78.9 & 87.5 & 59.9 & 63.5 \\
\bottomrule
\end{tabular}
\vspace{-4mm}
\end{table}

\subsection{Ablation Study}

Table~\ref{tab:multi_scale_ablation} reports the ablation results of HIPIF across different architectures and model scales. 
Here, \textbf{w/o Reflection} removes the hierarchical reflection mechanism, 
\textbf{w/o Reward} removes the subgoal-oriented process rewards and only uses the final task success signal for training, 
and \textbf{w/o Subgoal} removes the explicit subgoal decomposition and context folding structure. 

\textbf{Ablation on model architectures.} From Table~\ref{tab:multi_scale_ablation}, we could conclude that: (1) Removing the subgoal structure(w/o Subgoal) leads to the most significant performance drop, particularly on complex tasks such as PICK2, indicating that subgoal-level planning and context folding are fundamental to HIPIF. (2) Removing hierarchical reflection(w/o Reflection) mechanism substantially degrades performance, showing that hierarchical reflection helps the agent assess subgoal progress and correct subgoal-level failures without costly expert trajectories or auxiliary models. (3) Removing subgoal-oriented process rewards (w/o Reward) results in noticeable performance drops, suggesting that final task-level feedback alone provides limited guidance for reliable subgoal content and subgoal execution.

\textbf{Analysis across model scales.}
Furthermore, we examine the effect of model scale using 3B and 7B backbones. As shown in Table~\ref{tab:multi_scale_ablation}, HIPIF consistently achieves the best results under both settings, indicating that subgoal-centric training remains effective as the base model becomes stronger. 
Notably, the 3B HIPIF already outperforms the 7B variant without subgoal structure across all three benchmarks. 
This suggests that explicitly learning the organization of the subgoal is more critical than simply increasing the size of the model for long-horizon decision-making.

\subsection{Analysis on Efficiency}

\begin{wrapfigure}{r}{0.40\textwidth}
    \vspace{-1.4em}
    \centering

    \begin{minipage}{\linewidth}
    \captionsetup{type=table, width=\linewidth, justification=centering, singlelinecheck=false, font=small, skip=1pt}
    \caption{Token efficiency comparison.}
    \label{tab:token_efficiency}
    \centering
    \tiny
    \setlength{\tabcolsep}{1.5pt}
    \renewcommand{\arraystretch}{0.95}
    \resizebox{\linewidth}{!}{
    \begin{tabular}{lcccccc}
    \toprule
    \multirow{2}{*}{Method}
    & \multicolumn{2}{c}{ALFWorld $\downarrow$}
    & \multicolumn{2}{c}{VirtualHome $\downarrow$}
    & \multicolumn{2}{c}{ScienceWorld $\downarrow$} \\
    \cmidrule(lr){2-3}
    \cmidrule(lr){4-5}
    \cmidrule(lr){6-7}
    & Steps & Tok.(k) & Steps & Tok.(k) & Steps & Tok.(k) \\
    \midrule
    HIPIF & \textbf{16.5} & \textbf{16.6} & \textbf{29.5} & \textbf{33.5} & \textbf{15.0} & \textbf{22.0} \\
    - w/o Subgoal & 38.0 & 37.8 & 37.1 & 42.8 & 19.1 & 28.7 \\
    - w/o Reflection & 19.6 & 18.1 & 31.6 & 35.6 & 16.6 & 22.2 \\
    - w/o Reward & 20.7 & 20.3 & 33.2 & 37.8 & 17.8 & 26.1 \\
    GIGPO & 20.7 & 21.7 & 30.4 & 35.1 & 19.8 & 37.6 \\
    Hiagent+GRPO & 25.8 & 26.8 & 36.2 & 40.8 & 16.7 & 23.5 \\
    \bottomrule
    \end{tabular}
    }
    \end{minipage}

    \vspace{0.25em}

    \begin{minipage}{\linewidth}
    \captionsetup{type=figure, width=\linewidth, justification=centering, singlelinecheck=false, font=small, skip=1pt}
    \centering
    \includegraphics[width=0.96\linewidth]{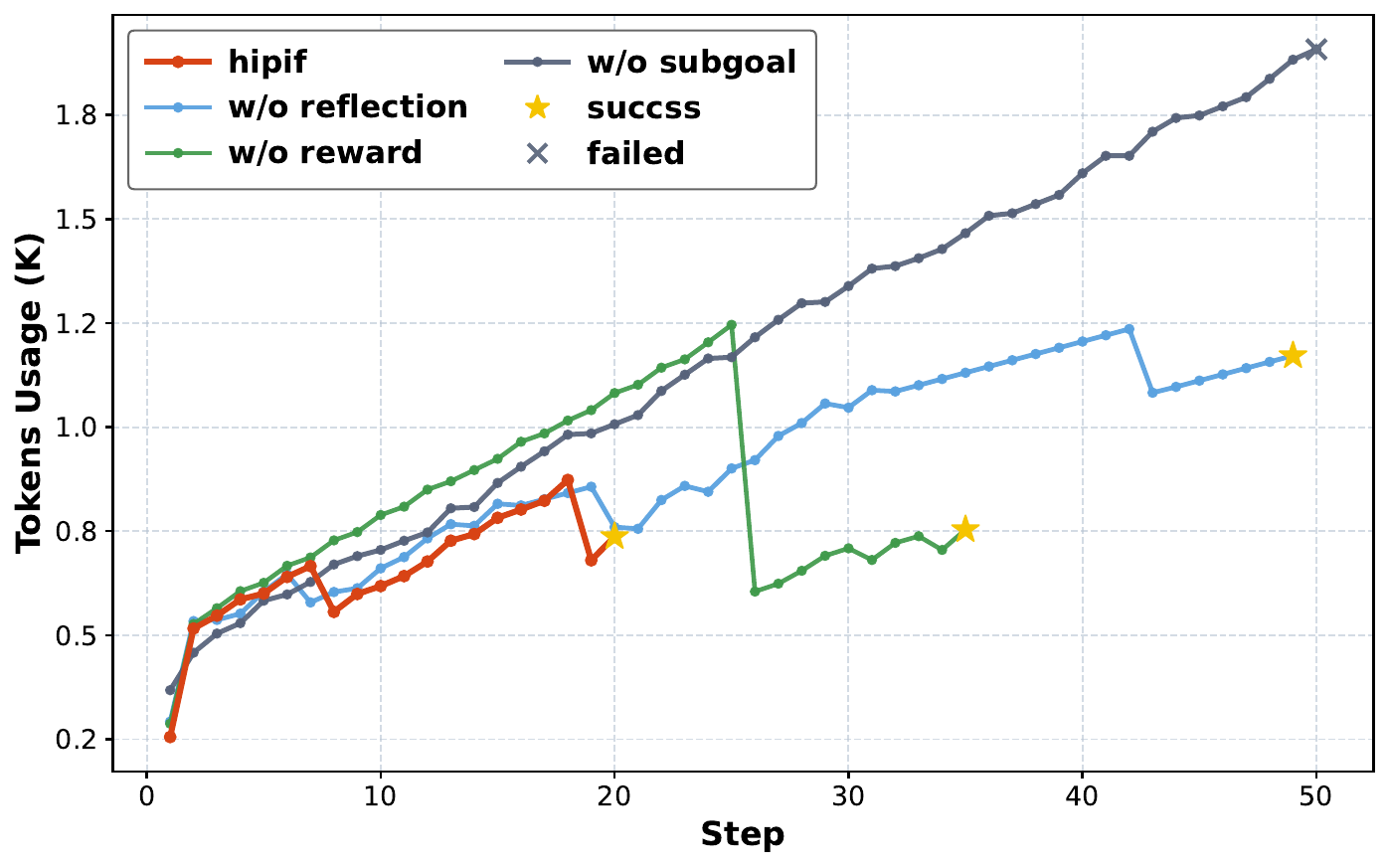}
    \caption{Per-step token consumption.}
    \label{fig:token_efficiency}
    \end{minipage}

    \vspace{-1.4em}
\end{wrapfigure}
\textbf{Token Efficiency.}
Table~\ref{tab:token_efficiency} evaluates context efficiency using two metrics: average completion steps and average input tokens per trajectory. HIPIF achieves the lowest cost across all three benchmarks, showing that it reduces both the number of interaction steps and the accumulated context tokens.
Figure~\ref{fig:token_efficiency} further compares the per-step token consumption of HIPIF and its ablated variants on the same ALFWorld task. We could observe that: (1) - w/o Subgoal keeps accumulating tokens as the interaction history grows, whereas HIPIF compresses completed subgoals into compact states, leading to lower token usage after each subgoal is finished. (2) Although reflection slightly increases original token usage, it helps the model better judge whether a subgoal has been completed, thereby avoiding unnecessary subsequent steps. 
(3) - w/o Reward receives weaker guidance on how to execute each subgoal, making it more likely to repeat ineffective actions and consume more tokens within a subgoal.

\begin{wrapfigure}{r}{0.35\textwidth}
    \vspace{-1.0em}
    \centering

    \begin{minipage}{\linewidth}
    \captionsetup{
        type=table,
        width=\linewidth,
        justification=centering,
        singlelinecheck=false,
        font=small,
        skip=1pt
    }
    \caption{Pipeline Efficiency.}
    \label{tab:efficiency_compare}
    \centering
    \tiny
    \setlength{\tabcolsep}{1.6pt}
    \renewcommand{\arraystretch}{0.92}
    \resizebox{0.96\linewidth}{!}{%
    \begin{tabular}{@{}lcc@{}}
    \toprule
    Method & Expert Traj. & Additional Models \\
    \midrule
    GRPO & \xmark & \xmark \\
    HiPER & \xmark & \cmark~critic \\
    GLIDER & \cmark~$2\times$ & \cmark~1 auxiliary model \\
    STEP-HRL & \cmark~$3\times$ & \cmark~2 auxiliary models \\
    HIPIF & \xmark & \xmark \\
    \bottomrule
    \end{tabular}
    }
    \end{minipage}

    \vspace{-0.8em}
\end{wrapfigure}

\textbf{Pipeline Efficiency.}
We further compare the pipeline efficiency of different methods. As shown in Table~\ref{tab:efficiency_compare}, HIPIF and GRPO are lightweight, requiring neither task-specific expert trajectories nor additional models. 
By contrast, HiPER needs an extra critic model, while GLIDER and STEP-HRL rely on task-specific expert trajectories and auxiliary models for subgoal generation or context compression. 
Together, Table~\ref{tab:evaluation_results} and Table~\ref{tab:efficiency_compare} show that HIPIF achieves strong effectiveness, high efficiency, and environmental scalability.

\begin{table}[htbp]
\captionsetup{justification=raggedright,singlelinecheck=false}
\centering
\caption{Case study on a long-horizon PICK2 task in ALFWorld, illustrating the effect of HIPIF.}
\label{tab:case_study}
\vspace{1mm}
\scriptsize
\setlength{\tabcolsep}{1.8pt}
\renewcommand{\arraystretch}{1.02}
\resizebox{\textwidth}{!}{%
\begin{tabular}{L{1.0cm}L{4.3cm}L{4.3cm}L{4.3cm}}
\toprule
\textbf{Method} 
& \multicolumn{1}{c}{\textbf{HIPIF}} 
& \multicolumn{1}{c}{\textbf{w/o Subgoal}} 
& \multicolumn{1}{c}{\textbf{w/o Reflection}} \\
\midrule
\textbf{Task}
&
\multicolumn{3}{L{13.5cm}}{
 Welcome to TextWorld, ALFRED! =- You are in the middle of a room. Looking quickly around you, you see a bathtubbasin 1, a drawer 1, a cabinet 3, a cabinet 2, a cabinet 1, a countertop 1, a garbagecan 1, a handtowelholder 1, a sinkbasin 2, a sinkbasin 1, a toilet 1, a toiletpaperhanger 1, and a towelholder 1. Your task is to: put two toiletpaper in toilet
}
\\
\midrule
\textbf{History}
&
\textbf{Subgoal 1}: find the first toiletpaper. 
Res 1: On drawer 1, you see toiletpaper 1.
\newline
\textbf{Subgoal 2}: take the first toiletpaper from drawer 1.
Res 2: Toiletpaper 1 is picked up from drawer 1.
\newline
\textbf{Subgoal 3}: put the first toiletpaper in toilet.
Res 3: Toiletpaper 1 is moved to toilet 1.
\newline
\textbf{Current\_Subgoal}: find the second toiletpaper.
Action: go to countertop 1.
Obs: On the countertop 1, you see a candle 1, and a cloth 2. 
\par
\textcolor{omitgray}{\textbf{[... omitted following steps ...]}}
\par
&
\textbf{Action 1}: go to drawer 1. 
Obs 1: Toiletpaper 1 are observed on drawer 1.
\newline
\textbf{Action 2}: take toiletpaper 1 from drawer 1.
Obs 2: Toiletpaper 1 is picked up.
\newline
\textbf{Action 4}: take toiletpaper 1 from inventory.
Obs 4: Nothing happens.
\par
\textcolor{omitgray}{\textbf{[... omitted intermediate steps ...]}}
\par
\textbf{Action 24}: move toiletpaper 1 to toilet 1. 
Obs 24: Nothing happens.
\newline
\textbf{Action 25}: take toiletpaper 1 from inventory. 
Obs 25: Nothing happens.
\par
\textcolor{omitgray}{\textbf{[... omitted following steps ...]}}
\par
&
\textbf{Subgoal 1}: find the first toiletpaper. 
Res 1: On drawer 1, you see toiletpaper 1.
\newline
\textbf{Subgoal 2}: take the first toiletpaper from drawer 1.
Res 2: Toiletpaper 1 is picked up from drawer 1.
\newline
\textbf{Subgoal 3}: put the first toiletpaper in toilet.
Res 3: Toiletpaper 1 is moved to toilet 1.
\newline
\textbf{Current\_Subgoal}: find the second toiletpaper.
Action: go to countertop 1.
Obs: On the countertop 1, you see a candle 1, and a cloth 2. 
\par
\textcolor{omitgray}{\textbf{[... omitted following steps ...]}}
\par
\\

\midrule

\textbf{Key step}
&
\textcolor{successgreen}{\textbf{Reflection:} Part1: The observation shows that on the countertop 1, there is no mention of any toiletpaper. The current subgoal is not completed. Part2: The next action should be to go to the toiletpaperhanger 1 to check if there is any toiletpaper.
\newline
\textbf{Action:} go to toiletpaperhanger 1.}
\newline
\textbf{Observation:} On the toiletpaperhanger 1, you see toiletpaper 4.
&
\textcolor{failred}{\textbf{Thinking:} Based on the previous observations, I have already token toiletpaper 1 from drawer 1. I can take the second piece of toiletpaper from the countertop 2 and put it in the toilet 1. 
\newline
\textbf{Action:} move toiletpaper 1 to toilet 1.}
\newline
\textbf{Observation:} Nothing happens.
&
\textcolor{failred}{\textbf{Reflection:} I have successfully found and put one toiletpaper in the toilet. Now, I need to put the second piece of toiletpaper in the toilet. The current subgoal is to put the second toiletpaper in the toilet.  
\newline
\textbf{Subgoal:} Put the second toiletpaper in the toilet.
\newline
\textbf{Action:} put toiletpaper 2 on toilet 1.}
\newline
\textbf{Observation:} Nothing happens.
\\
\midrule

\textbf{Outcome}
&
\textcolor{successgreen}{\textbf{Success.}}
&
\textcolor{failred}{\textbf{Failure.}}
&
\textcolor{failred}{\textbf{Failure.}}
\\

\bottomrule
\end{tabular}
}
\end{table}

\subsection{Case Study}

We conduct a case study on ALFWorld to illustrate the effect of our core designs, as shown in Table~\ref{tab:case_study}. Concretely, we observe that: (1) Subgoal decomposition and context folding help HIPIF maintain a clear task structure under long contexts. After placing the first toiletpaper, HIPIF correctly identifies the current stage as finding the second one and continues planning under this subgoal. In contrast, \textbf{w/o Subgoal} reasons over an unstructured long action history, loses track of the completed progress, and incorrectly returns to the already handled toiletpaper 1 with invalid actions. 
(2) Hierarchical reflection enables reliable subgoal progress assessment and transition. HIPIF determines from the current observation that the second toiletpaper has not been found, and therefore continues the search instead of switching stages. 
By contrast, \textbf{w/o Reflection} keeps the subgoal history but lacks explicit completion checking, causing a premature transition to the next subgoal and eventual failure. Additional case studies on ScienceWorld and VirtualHome are presented in Appendix~\ref{app:case_studies_addition}.

\section{Conclusions}

In this paper, we proposed \textbf{Hi}erarchical \textbf{P}lanning and \textbf{I}nformation \textbf{F}olding (HIPIF) to improve long-horizon decision-making for LLM agents without relying on task-specific expert trajectories or auxiliary models. HIPIF trains the agent end-to-end to plan explicit subgoals and fold completed execution of subgoals, thus reducing long-context interference. To ensure reliable subgoal-based execution, HIPIF further introduces hierarchical reflection and subgoal-oriented process rewards to guide subgoal assessment, transition, and execution. Empirical results on three publicly available benchmarks demonstrate that HIPIF consistently outperforms other methods while improving efficiency. We believe that hierarchical planning and information folding offers a promising direction for improving both effectiveness and efficiency in future agentic decision-making research.

\bibliographystyle{plainnat}
\bibliography{references}


\appendix

\section{Datasets Details}
\label{app:datasets}

\paragraph{ALFWorld.}
ALFWorld~\citep{shridharalfworld} is an embodied text-based environment designed to evaluate the agentic capability of language models in complex long-horizon decision-making tasks. 
It provides interactive TextWorld environments that are closely aligned with ALFRED~\citep{shridhar2020alfred}. 
In each episode, the agent receives a textual goal and interacts with the environment through multi-turn action generation until the task is completed or the maximum number of interaction turns is reached. 
The environment returns a binary outcome as the trajectory-level score, where $1$ denotes task success and $0$ denotes failure. 
ALFWorld contains six categories of common household tasks: Pick \& Place (\textsc{Pick}), Examine in Light (\textsc{Look}), Clean \& Place (\textsc{Clean}), Heat \& Place (\textsc{Heat}), Cool \& Place (\textsc{Cool}), and Pick Two \& Place (\textsc{Pick2}). 
Following prior work, we adopt the version constructed by \citet{song2024trial} and set the maximum number of interaction turns to 50.

\paragraph{VirtualHome.}
VirtualHome~\citep{puig2018virtualhome} is another embodied household environment for evaluating long-horizon interactive agents. 
It includes diverse high-level household tasks in simulated indoor environments, where each task requires the agent to complete a sequence of executable actions. 
For each episode, the agent receives a high-level task description, then repeatedly selects an action and receives environment feedback until the task succeeds or the maximum interaction limit is reached. 
We use binary success as the evaluation outcome, where $1$ indicates successful task completion and $0$ indicates failure. 
In this work, we adopt the version provided by \citet{wang2025spa} and further correct several erroneous task descriptions in the original benchmark. 
The processed dataset will be released together with our code. 
We set the maximum number of interaction turns to 50.

\paragraph{ScienceWorld.}
ScienceWorld~\citep{wang2022scienceworld} is a text-based interactive environment designed to evaluate agents on scientific reasoning and experimental tasks. 
ScienceWorld requires the agent to perform multi-step scientific procedures, such as finding relevant objects,  observing physical or chemical changes, and so on. 
At each step, the agent generates an executable textual action and receives an observation from the environment. 
The interaction continues until the task is completed or the maximum number of interaction turns is reached. 
For consistency with the other benchmarks, we report binary success rates, where $1$ denotes task success and $0$ denotes failure. In this work, we adopt the version provided by STEP-HRL~\citep{zhen2026hierarchical}. We set the maximum number of interaction turns to 40.

To ensure a fair comparison, we follow the dataset settings of the strongest baseline GIGPO~\citep{fenggroup} and STEP-HRL~\citep{zhen2026hierarchical}. 
Table~\ref{tab:dataset_stats} summarizes the statistics of the three benchmarks used in our experiments.

\begin{table}[htbp]
\centering
\caption{Statistics of the three agent benchmarks. Train and Test denote the numbers of training and test samples. Available Actions denotes the number of action templates used by the agent.}
\label{tab:dataset_stats}
\vspace{1mm}
\small
\setlength{\tabcolsep}{8pt}
\renewcommand{\arraystretch}{1.05}
\begin{tabular}{lcccc}
\toprule
Benchmark & Train & Test & Available Actions & Maximum Turns \\
\midrule
ALFWorld & 2,851 & 134 & 13 & 50 \\
VirtualHome & 4,920 & 247 & 40 & 50 \\
ScienceWorld & 1483 & 211 & 17 & 40 \\
\bottomrule
\end{tabular}
\end{table}

\section{Training Details}
\label{app:training_details}

\paragraph{Hyperparameters.}
For all three benchmarks, including ALFWorld, VirtualHome, and ScienceWorld, we use the same training configuration unless otherwise specified. 
The maximum prompt length is set to 8192 tokens, and the maximum response length is set to 512 tokens. 
The actor learning rate is set to $1\times10^{-6}$. 
For PPO, which is the only method that uses a critic model, the critic learning rate is set to $1\times10^{-5}$.
For group-based RL methods, including GRPO and its variants, we use a group size of 8 and sample 16 groups per rollout, resulting in $16 \times 8 = 128$ parallel environments. 
For PPO, we use 128 independent environments for rollout collection. 
The rollout temperature is set to 1.0 for exploration, while the validation temperature is set to 0.4 for more stable evaluation. 
The mini-batch size is set to 256, and the KL-divergence loss coefficient is set to 0.01. 

\paragraph{Reward Design.}
All methods use rule-based environment rewards. 
The terminal reward is 10 for task success and 0 for failure. We further assign a penalty of $-0.1$ when the model fails to produce the required structured tags, such as \texttt{<reflection>...</reflection>}, \texttt{<subgoal>...</subgoal>}, or \texttt{<action>...</action>}.

For HIPIF, we further incorporate subgoal-oriented process rewards to provide step-level feedback for subgoal content and subgoal execution. 
\textbf{For subgoal content}, we introduce two rule-based penalties. 
First, we extract object and location names from the generated subgoal and check whether they can be matched by string matching to entities in the available environment context.
If the subgoal contains no grounded object or location, we assign a penalty of $-0.1$. 
Second, for successful trajectories, we inspect the final observation of each terminated subgoal. 
If the final observation indicates an execution failure, such as ``Nothing happens'' or ``No known action matches that input'', we assign a penalty of $-0.1$ to the corresponding subgoal step. 
This helps identify locally erroneous subgoals that may be hidden inside an otherwise successful trajectory.
\textbf{For subgoal execution}, we penalize repeated ineffective interaction patterns within the same subgoal. 
Specifically, we compare the action-observation records under the current subgoal; if the same action-observation pair appears for the third time, we assign a penalty of $-0.1$. 
This rule is applied only within the temporal span of the same subgoal, since the same action may still be valid in different task stages. 

\paragraph{Computing Details.}
All training experiments are conducted on 8 NVIDIA A100 GPUs. 
For 7B models, we set the tensor parallel size to 4 and train for 150 epochs. 
For 3B models, we set the tensor parallel size to 2 and train for 200 epochs. 

\section{ Pseudo Code}
\label{app:pseudo}

Algorithm~\ref{alg:hipif} summarizes the overall training procedure of HIPIF. Here, $z_t$ denotes the completion judgment of the current subgoal, and $\eta_t$ denotes the rationale for this judgment. 
$\xi_t$ is the branch-specific reflection generated before the next decision: it reflects on subgoal transition when $z_t=completed$, and on current subgoal execution when $z_t=uncompleted$. 
The process reward $r_t^{\mathrm{proc}}$ is computed from subgoal-content and subgoal-execution feedback, and $Q_t$ denotes the corresponding step-level return used for GRPO optimization.

For each task instruction, we sample a group of trajectories from the old policy and perform rollout under the current folded-history structure. 
At each interaction step, the model first conducts hierarchical reflection and outputs a completion judgment $z_t$ together with a rationale $\eta_t$. 
This judgment determines the subsequent generation branch. 
If the current subgoal is judged as completed, HIPIF folds the completed subgoal and its intra-subgoal execution history into the compact history $\bar{H}$, appends the subgoal to the historical subgoal sequence $\mathcal{G}$, and then generates a reflection $\xi_t$ on the next subgoal together with the next subgoal and its first action. 
If the current subgoal is not completed, the model instead generates an execution reflection $\xi_t$ based on the current subgoal, recent intra-subgoal history, and latest observation, and then outputs the next action under the current subgoal.
Importantly, the reflection $\xi_t$ and the structured decision output $y_t$ are treated as part of the same policy-generated sequence. 
After rollout, HIPIF computes the terminal environment reward and the subgoal-oriented process rewards for each stored step, obtains step-level returns, normalizes them within the sampled group, and updates the policy with the clipped GRPO objective.

\begin{algorithm}[t]
\small
\caption{Training Procedure of HIPIF}
\label{alg:hipif}
\begin{algorithmic}[1]
\Require Training tasks $\mathcal{D}$, policy $\pi_\theta$, old policy $\pi_{\theta_{\mathrm{old}}}$, group size $M$, horizon $T$
\Ensure Updated policy $\pi_\theta$

\For{each training iteration}
    \For{each task instruction $q \in \mathcal{D}$}
        \State $\mathcal{Y} \leftarrow \emptyset$
        \For{$m=1$ to $M$}
            \State Reset environment with $q$ and obtain initial observation $o_0$
            \State Initialize folded history $\bar{H}\leftarrow\emptyset$, subgoal sequence $\mathcal{G}\leftarrow\emptyset$
            \State Generate initial subgoal $g$ and set intra-subgoal history $h\leftarrow\emptyset$
            
            \For{$t=0$ to $T-1$}
                \State $(z_t,\eta_t)\leftarrow \textsc{Reflect}(\pi_{\theta_{\mathrm{old}}}, \bar{H}, \mathcal{G}, g, h, o_t)$
                
                \If{$z_t=completed$}
                    \State $\bar{H}\leftarrow \textsc{Fold}(\bar{H}, g, h, o_t)$, \quad $\mathcal{G}\leftarrow \mathcal{G}\Vert g$
                    \State $(\xi_t, y_t)\leftarrow \textsc{Generate}(\pi_{\theta_{\mathrm{old}}}, \bar{H}, \mathcal{G}, \eta_t)$
                    \State Parse $y_t$ as the next subgoal $g$ and its first action $a_t$
                    \State Reset intra-subgoal history $h\leftarrow\emptyset$
                \Else
                    \State $(\xi_t, y_t)\leftarrow \textsc{Generate}(\pi_{\theta_{\mathrm{old}}}, g, h, o_t, \eta_t)$
                    \State Parse $y_t$ as the next action $a_t$ under current subgoal $g$
                \EndIf
                
                \State Validate the output schema, subgoal grounding, and action validity
                \State Execute $a_t$, receive $o_{t+1}$, and append $(a_t,o_{t+1})$ to $h$
                
                \If{task is completed}
                    \State \textbf{break}
                \EndIf
            \EndFor
            
            \State Obtain terminal reward $R_{\mathrm{env}}^{(m)}$
            \State Compute process reward $r_{t}^{\mathrm{proc},(m)}$ for each stored step
            \State Compute step-level return $Q_t^{(m)}=R_{\mathrm{env}}^{(m)}+r_t^{\mathrm{proc},(m)}$
            \State Add trajectory $m$ to $\mathcal{Y}$
        \EndFor
        
        \State Normalize $\{Q_t^{(m)}\}$ within $\mathcal{Y}$ to obtain step-level advantages $\hat{A}_{t}^{(m)}$
        \State Update $\pi_\theta$ with the clipped GRPO objective
    \EndFor
\EndFor

\end{algorithmic}
\end{algorithm}

\section{Sensitivity Analysis}
\label{app:sensitivity_analysis}

\begin{figure*}[t]
\centering
\begin{subfigure}[b]{0.48\textwidth}
    \centering
    \includegraphics[width=\textwidth]{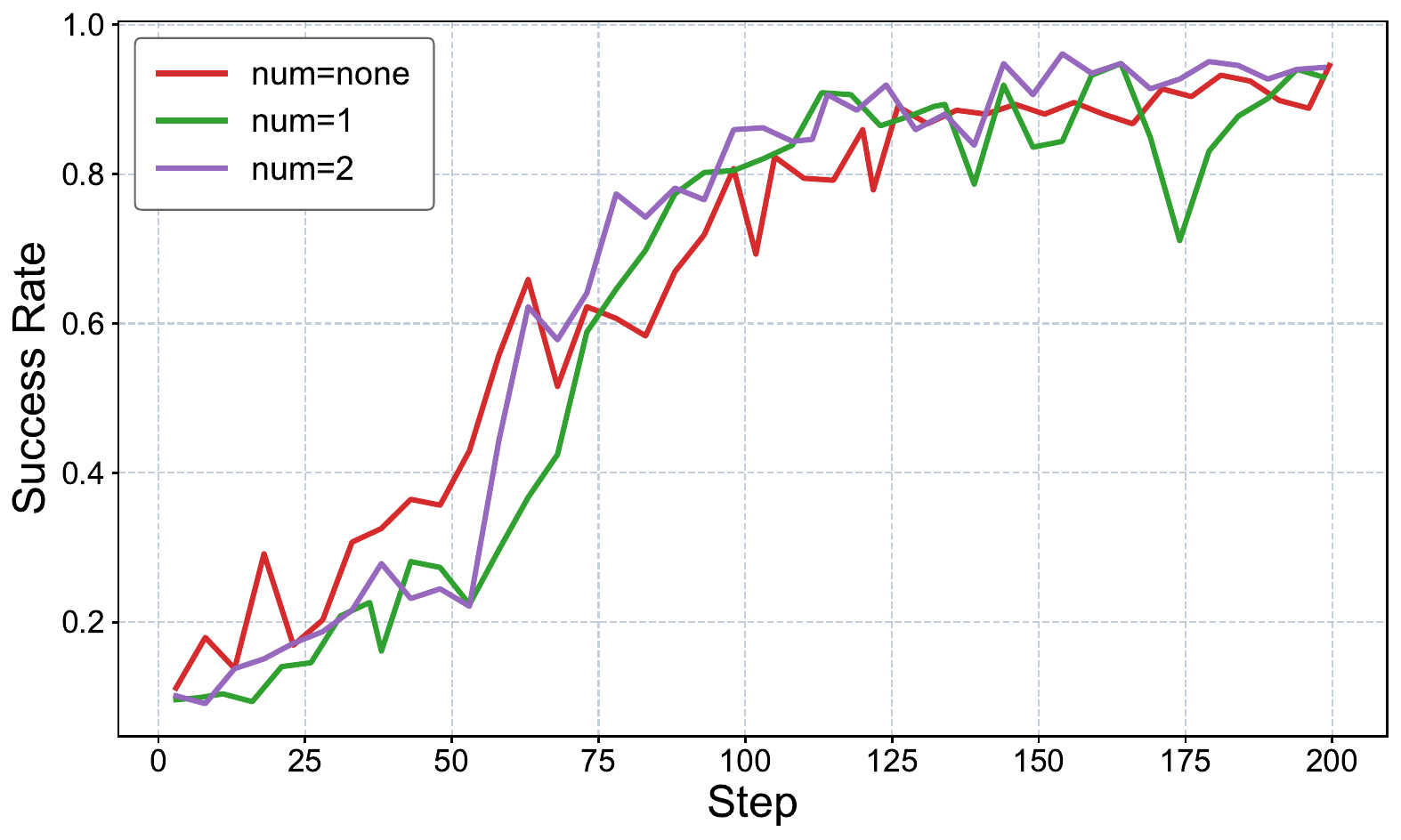}
    \caption{Sensitivity to repetition threshold.}
    \label{fig:sensitivity_action}
\end{subfigure}
\hfill
\begin{subfigure}[b]{0.48\textwidth}
    \centering
    \includegraphics[width=\textwidth]{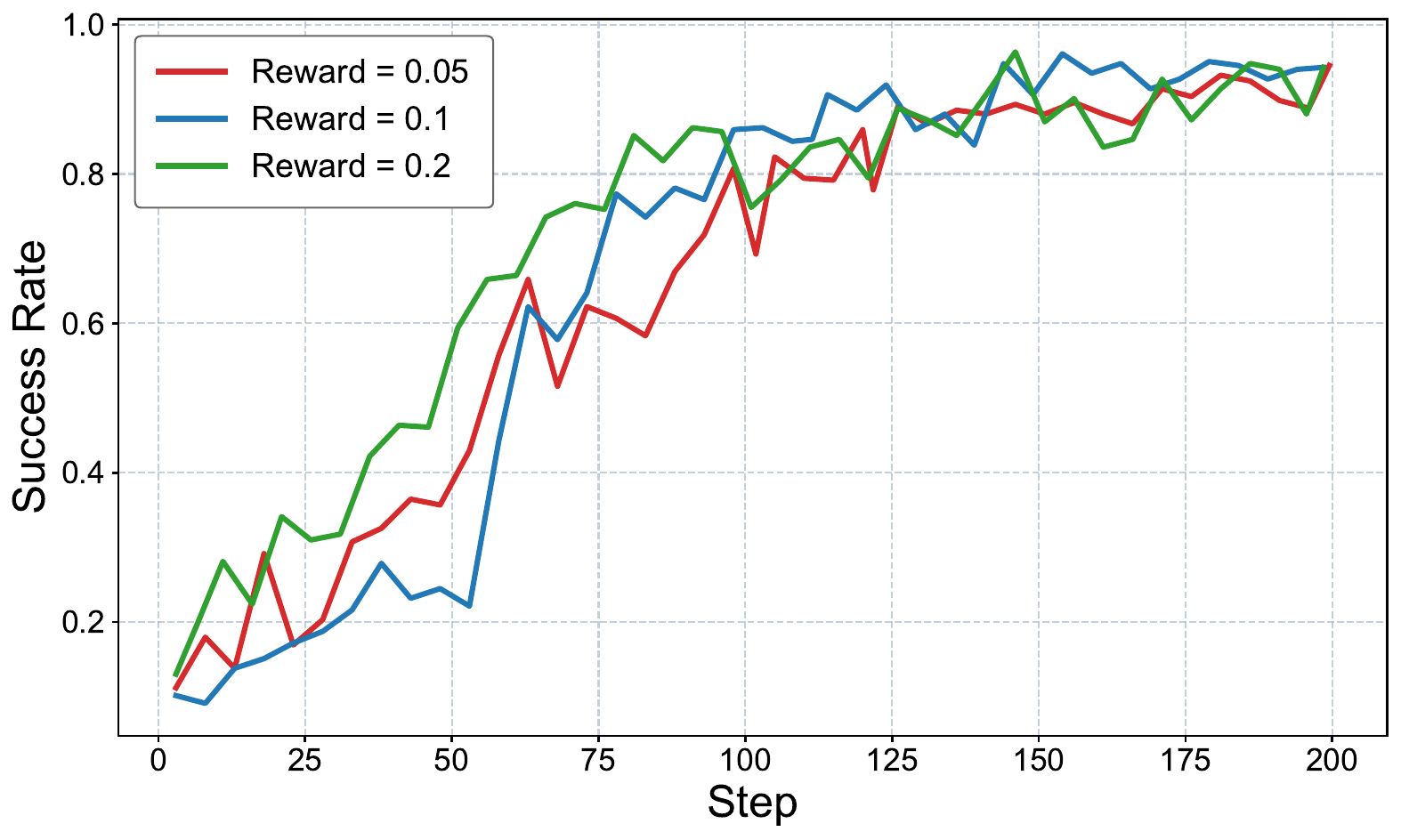}
    \caption{Sensitivity to reward magnitude.}
    \label{fig:sensitivity_reward}
\end{subfigure}
\caption{Sensitivity analysis of the subgoal-oriented process rewards on AlfWorld.}
\label{fig:sensitivity_analysis}
\end{figure*}

We conduct sensitivity analysis on two key hyperparameters in the subgoal-oriented process rewards: the repetition threshold for action-observation pairs and the magnitude of the process penalty. 
The experiments are conducted on  on AlfWorld and results are shown in Figure~\ref{fig:sensitivity_analysis}.

For the repetition penalty, we compare three settings: no penalty, penalizing the first repeated action-observation pair, and penalizing the second repeated occurrence. 
As shown in Figure~\ref{fig:sensitivity_action}, penalizing the first repeated pair leads to larger late-stage fluctuations. 
In contrast, penalizing the second repeated occurrence yields more stable performance throughout training.

For the penalty magnitude, we compare values of $0.05$, $0.1$, and $0.2$. 
As shown in Figure~\ref{fig:sensitivity_reward}, a larger penalty of $0.2$ improves early performance but causes clear fluctuations in later training, while $0.05$ provides weaker supervision and performs slightly worse. 
The value of $0.1$ achieves the best overall balance between effectiveness and stability.

\section{Templates}
\label{app:prompt_memory}

\paragraph{Prompt Templates.}
Figures~\ref{fig:prompt_alfworld}, \ref{fig:prompt_scienceworld}, and \ref{fig:prompt_virtualhome} present the prompt templates used for ALFWorld, ScienceWorld, and VirtualHome, respectively. 
To better match the interactive agent setting, we do not provide the model with a fully grounded list of executable actions at each step. 
Instead, we provide environment-specific action templates and require the model to instantiate valid objects or locations according to the current observation. 
This setting encourages the agent to explore the environment and learn executable behavior from interaction feedback, rather than relying on an external action enumerator.
These prompt templates are constructed using Python-style string formatting, where placeholders enclosed in curly braces, such as \texttt{\{task\_description\}}, \texttt{\{current\_subgoal\}}, \texttt{\{current\_observation\}}, and \texttt{\{action\_history\}}, denote semantic slots that are dynamically filled at runtime using Python's \texttt{.format()} function. 
Each prompt provides the task description, current subgoal, current observation, last action, available action templates, and folded execution history. 
The output format is also explicitly constrained by tags: the model first outputs its reflection process inside \texttt{<reflection>...</reflection>}, and then outputs either a new subgoal with its first action using \texttt{<subgoal>...</subgoal>} and \texttt{<action>...</action>}, or only the next action using \texttt{<action>...</action>} when the current subgoal should continue. 

\begin{figure}[htbp]
\centering
\begin{promptbox}{Prompt Template for ALFWorld}
\scriptsize

You are an expert agent operating in the ALFRED Embodied Environment. Your task is to: \pvar{task_description}\par
Your current subgoal is: \pvar{current_subgoal}.\par
Your current observation is: \pvar{current_observation}\par
Your last action is: \pvar{last_action}.\par
\vspace{0.4em}

\strong{Here are the AVAILABLE ACTIONS you could take:}\par
\texttt{- 'go to \{\{recep\}\}'}\par
\texttt{- 'take \{\{obj\}\} from \{\{recep\}\}'}\par
\texttt{- 'put \{\{obj\}\} on \{\{recep\}\}'}\par
\texttt{- 'open \{\{recep\}\}' / 'close \{\{recep\}\}'}\par
\texttt{- 'use \{\{obj\}\}/\{\{recep\}\}'}\par
\texttt{- 'clean \{\{obj\}\} with \{\{recep\}\}'}\par
\texttt{- 'heat \{\{obj\}\} with \{\{recep\}\}'}\par
\texttt{- 'cool \{\{obj\}\} with \{\{recep\}\}'}\par
Ensure that any objects \texttt{('\{\{obj\}\}')} and receptacles \texttt{('\{\{recep\}\}')} are present in your observation. The current observation \texttt{"Nothing happens."} means the action failed.\par
\vspace{0.4em}

\strong{EXECUTION HISTORY:}\par
Subgoals are essential milestones on the path to the final task. Below is the execution history, each entry enclosed in \texttt{[]} contains a subgoal followed by all corresponding actions and observations: \pvar{action_history}\par
\par
\vspace{0.4em}

\strong{REFLECTION:}\par
You should first output your reflection process in the following format: \texttt{<reflection>the reflection process</reflection>}.\par
The reflection process includes two parts:\par
\texttt{- Part1:} Output whether the current subgoal is completed or not according to the current observation.\par
\texttt{- Part2:} If the current subgoal is completed, output your reflection on what the new subgoal should be to advance the final task based on historical subgoals. If the current subgoal is not completed, output your reflection on the next action based on the execution of the current subgoal.\par
\vspace{0.4em}

\strong{ACTION:}\par
Once you have finished the reflection process, choose one output format based on whether the current subgoal is completed:\par
\texttt{- If the current subgoal is completed:} output a new subgoal and its first action in the format \texttt{<subgoal>your subgoal</subgoal> <action>your action</action>}.\par
\texttt{- If the current subgoal is not completed:} output the next action in the format \texttt{<action>your action</action>}.\par

\end{promptbox}
\caption{Prompt template for ALFWorld.}
\label{fig:prompt_alfworld}
\end{figure}

\begin{figure}[htbp]
\centering
\label{fig:prompt_virtualhome}
\label{fig:actor_prompt_virtualhome}
\begin{promptbox}{Prompt Template for ScienceWorld}
\scriptsize

You are a helpful assistant to do some scientific experiment in an environment.\par
Your current subgoal is: \pvar{current_subgoal}.\par
Your last action is: \pvar{last_action}.\par
Your current observation is: \pvar{current_observation}\par
\vspace{0.3em}

\pvar{task_description}\par
\vspace{0.3em}

In the environment, there are several rooms: kitchen, foundry, workshop, bathroom, outside, living room, bedroom, greenhouse, art studio, hallway. You can teleport to any room in one step.\par
\vspace{0.3em}

\strong{Here are the AVAILABLE ACTIONS you could take:}\par
\texttt{- 'open \{\{object\}\}' / 'close \{\{object\}\}': open or close a container}\par
\texttt{- 'connect \{\{object\}\} to \{\{object\}\}': connect electrical components}\par
\texttt{- 'pick up \{\{object\}\}': move an object to the inventory}\par
\texttt{- 'put down \{\{object\}\}': drop an inventory item}\par
\texttt{- 'move \{\{object\}\} to \{\{object\}\}': move an object to a container}\par
\texttt{- 'pour \{\{object\}\} into \{\{object\}\}': pour a liquid into a container}\par
\texttt{- 'mix \{\{object\}\}': chemically mix a container}\par
\texttt{- 'use \{\{object\}\} on \{\{object\}\}': use thermometer on object to measure its temperature}\par
\texttt{- 'activate \{\{object\}\}' / 'deactivate \{\{object\}\}': activate or deactivate a device}\par
\texttt{- 'teleport to \{\{location\}\}': move to a new location}\par
\texttt{- 'focus on \{\{object\}\}': signal intent on a task object}\par
\texttt{- 'wait': wait for 10 steps}\par
\texttt{- 'wait1': wait for a step}\par
\vspace{0.3em}

Ensure that any \texttt{\{\{object\}\}} and \texttt{\{\{location\}\}} are present in your observation. Do NOT use \texttt{'focus on \{\{object\}\}'} to get details.\par
\vspace{0.3em}

\strong{EXECUTION HISTORY:}\par
Subgoals are essential milestones on the path to the final task. Below is the execution history, each entry enclosed in \texttt{[]} contains a subgoal followed by all corresponding actions and observations: \pvar{action_history}\par
\par
\vspace{0.3em}

\strong{REFLECTION:}\par
You should first output your reflection process in the following format: \tagname{<reflection>the reflection process</reflection>}.\par
The reflection process includes two parts:\par
\texttt{- Part1:} Output whether the current subgoal, NOT the task, is completed or not completed according to the current observation.\par
\texttt{- Part2:} If the current subgoal is completed, output your reflection on what the new subgoal should be to advance the final task. If the current subgoal is not completed, output your reflection on the next action.\par
\vspace{0.3em}

\strong{ACTION:}\par
Once you have finished the reflection process, choose one output format based on your reflection process:\par
\texttt{- If completed:} output a new subgoal and its first action in the format \tagname{<subgoal>your subgoal</subgoal> <action>your action</action>}.\par
\texttt{- If not completed:} output the next action in the format \tagname{<action>your action</action>}.\par

\end{promptbox}
\caption{prompt template for ScienceWorld.}
\label{fig:prompt_scienceworld}
\end{figure}

\begin{figure}[t]
\centering
\begin{promptbox}{Prompt Template for VirtualHome}
\scriptsize

You are an agent in a simulated household environment, tasked to assist with daily living activities and interactions. Your task is to: \pvar{task_description}\par
Your current subgoal is: \pvar{current_subgoal}.\par
Your current observation is: \pvar{current_observation}\par
Your last action is: \pvar{last_action}.\par
\vspace{0.3em}

\strong{Here are the AVAILABLE ACTIONS you could take:}\par
\texttt{- walk to \{\{obj\}\}}\par
\texttt{- find \{\{obj\}\}}\par
\texttt{- grab \{\{obj\}\}}\par
\texttt{- open \{\{obj\}\}}\par
\texttt{- close \{\{obj\}\}}\par
\texttt{- put \{\{obj\}\} on \{\{recep\}\}}\par
\texttt{- put \{\{obj\}\} in \{\{recep\}\}}\par
\texttt{- switch on \{\{obj\}\}}\par
\texttt{- switch off \{\{obj\}\}}\par
\texttt{- drink \{\{obj\}\}}\par
\texttt{- sit on \{\{obj\}\}}\par
\texttt{- lie on \{\{obj\}\}}\par
\texttt{- look at \{\{obj\}\}}\par
\texttt{- stand up}\par
\texttt{- watch \{\{obj\}\}}\par
\texttt{- wipe \{\{obj\}\}}\par
\texttt{- type on \{\{obj\}\}}\par
\texttt{- take off \{\{obj\}\}}\par
\texttt{- wash \{\{obj\}\}}\par
\texttt{- cut \{\{obj\}\}}\par
\texttt{- eat \{\{obj\}\}}\par
\texttt{- sleep}\par
\texttt{- wake up}\par
\texttt{- plug in \{\{obj\}\}}\par
\texttt{- plug out \{\{obj\}\}}\par
\texttt{- pour \{\{obj\}\} into \{\{recep\}\}}\par
\texttt{- turn to \{\{obj\}\}}\par
\vspace{0.3em}

Ensure that any objects \texttt{('\{\{obj\}\}')} and locations \texttt{('\{\{recep\}\}')} are present in your observation. You should walk to the object before taking action on it.\par
\vspace{0.3em}

\strong{EXECUTION HISTORY:}\par
Subgoals are essential milestones on the path to the final task. Below is the execution history, each entry enclosed in \texttt{[]} contains a subgoal followed by all corresponding actions and observations: \pvar{action_history}\par
\par
\vspace{0.3em}

\strong{REFLECTION:}\par
You should first output your reflection process in the following format: \tagname{<reflection>the reflection process</reflection>}.\par
The reflection process includes two parts:\par
\texttt{- Part1:} Output whether the current subgoal is completed or not according to the current observation.\par
\texttt{- Part2:} If the current subgoal is completed, output your reflection on what the new subgoal should be to advance the final task based on historical subgoals. If the current subgoal is not completed, output your reflection on the next action based on the execution of the current subgoal.\par
\vspace{0.3em}

\strong{ACTION:}\par
Once you have finished the reflection process, choose one output format based on whether the current subgoal is completed in Part1:\par
\texttt{- If completed:} output a new subgoal and its first action in the format \tagname{<subgoal>your subgoal</subgoal> <action>your action</action>}.\par
\texttt{- If not completed:} output the next action in the format \tagname{<action>your action</action>}.\par

\end{promptbox}
\caption{Prompt template for VirtualHome.}
\label{fig:prompt_virtualhome}
\end{figure}

\paragraph{Folded Memory Templates.}
Figures~\ref{fig:folded_memory_example}, \ref{fig:folded_memory_virtualhome}, and \ref{fig:folded_memory_scienceworld} show folded subgoal-level memory examples for ALFWorld, VirtualHome, and ScienceWorld, respectively.
The memory is organized to support both global task tracking and local subgoal execution. 
Specifically, we first present the current subgoal memory, which contains the active subgoal and its recent action-observation records. 
We then include the initial environment information, which provides the starting state and task context. 
Finally, we append the historical folded memory, where each completed subgoal is stored as a compact record.
Each memory unit is enclosed in square brackets \texttt{[...]} to make different subgoals easy to distinguish. 
Within each memory unit, action-observation pairs are enclosed in parentheses, such as \texttt{(pre\_action: ..., pre\_observation: ...)}.

\begin{figure}[htbp]
\centering
\begin{memorybox}{Folded History Example on ALFWorlD}
\scriptsize

\textbf{Your task is to:} put a cool tomato in microwave.\par
\textbf{Your current subgoal is:} put the cooled tomato in the microwave.\par
\vspace{0.4em}

\strong{EXECUTION HISTORY:}\par
Subgoals are essential milestones on the path to the final task. Below is the execution history, each entry enclosed in \texttt{[]} contains a subgoal followed by all corresponding actions and observations:\par
\vspace{0.3em}

{[(\textbf{current\_subgoal}: put the cooled tomato in the microwave),}\par
\texttt{(pre\_action: go to microwave 1, pre\_observation: You arrive at microwave 1. The microwave 1 is closed.),}\par
\texttt{(pre\_action: open microwave 1, observation: You open the microwave 1. The microwave 1 is open. In it, you see a egg 1.)]}\par
\vspace{0.3em}

{[(\textbf{origin\_observation}: -= Welcome to TextWorld, ALFRED! =- You are in the middle of a room. Looking quickly around you, you see a cabinet 6, a cabinet 5, a cabinet 4, a cabinet 3, a cabinet 2, a cabinet 1, a coffeemachine 1, a countertop 3, a countertop 2, a countertop 1, a drawer 3, a drawer 2, a drawer 1, a fridge 1, a garbagecan 1, a microwave 1, a shelf 3, a shelf 2, a shelf 1, a sinkbasin 1, a stoveburner 4, a stoveburner 3, a stoveburner 2, a stoveburner 1, and a toaster 1. Your task is to: put a cool tomato in microwave.)]}\par
\vspace{0.3em}

{[(\textbf{history\_subgoal 1}: find a tomato),}\par
{(pre\_action: go to countertop 2, pre\_observation: You arrive at countertop 2. On the countertop 2, you see a butterknife 1, a cellphone 1, a creditcard 1, a knife 1, a lettuce 1, a saltshaker 2, a saltshaker 1, a statue 1, and a tomato 1.)]}\par
\vspace{0.3em}

{[(\textbf{history\_subgoal 2}: cool the tomato),}\par
{(pre\_action: cool tomato 1 with fridge 1, pre\_observation: You cool the tomato 1 using the fridge 1.)]}\par
\vspace{0.3em}

\end{memorybox}
\caption{Example of the folded subgoal-level memory used by HIPIF.}
\label{fig:folded_memory_example}
\end{figure}

\begin{figure}[htbp]
\centering
\begin{memorybox}{Folded History Example on VirtualHome}
\scriptsize

\textbf{Your task is to:} Wipe down counter. 
Instructions: Walk to dining room. Grab the rag, rinse it in the sink using the faucet, and then wipe the kitchen counter.\par
\textbf{Your current subgoal is:} open the faucet.\par
\vspace{0.4em}

\strong{EXECUTION HISTORY:}\par
Subgoals are essential milestones on the path to the final task. Below is the execution history, each entry enclosed in \texttt{[]} contains a subgoal followed by all corresponding actions and observations:\par
\vspace{0.3em}

{[(\textbf{current\_subgoal}: wipe kitchen\_counter),}\par
{(pre\_action: walk to kitchen\_counter, pre\_observation:you successfully walk to kitchen\_counter. kitchen\_counter is CLOSED and DIRTY, sink is CLEAN, faucet is OFF and CLEAN, rag is CLEAN, faucet ON sink, faucet ON kitchen\_counter. You have rag.)],}\par
\vspace{0.3em}

{[(\textbf{origin\_observation}: Available rooms in the house are: dining\_room, bedroom, home\_office, bathroom. You are currently in the bedroom.)]}\par
\vspace{0.3em}

{[(\textbf{history\_subgoal 1}: walk to dining room),}\par
{(pre\_action: walk to dining\_room, observation: you successfully walk to dining\_room. You arrive in the dining\_room. You see bench, bookshelf, chair, coffe\_maker, cupboard, dishwasher, faucet, freezer, kitchen\_counter, microwave, oven, phone, rag, sink, stovefan, table, toaster, tray, tvstand.)]}\par
\vspace{0.3em}

{[(\textbf{history\_subgoal 2}: find and grab the rag),}\par
{(pre\_action: grab rag, observation: you successfully grab rag. character, rag is CLEAN. You have rag.)]}\par
\vspace{0.3em}

{[(\textbf{history\_subgoal 3}: find the sink),}\par
{(pre\_action: walk to sink, observation: you successfully walk to sink. kitchen\_counter is CLOSED and CLEAN, sink is CLEAN, faucet is OFF and CLEAN, rag is CLEAN, faucet ON sink, faucet ON kitchen\_counter. You have rag.)]}\par
\vspace{0.3em}

{[(\textbf{history\_subgoal 4}: find the faucet),}\par
{(pre\_action: find faucet, observation: you successfully find faucet. kitchen\_counter is CLOSED and CLEAN, sink is CLEAN, faucet is OFF and CLEAN, rag is CLEAN, faucet ON sink, faucet ON kitchen\_counter. You have rag.)]}\par
\vspace{0.3em}

{[(\textbf{history\_subgoal 5}: rinse the rag in the sink using the faucet),}\par
{(pre\_action: Rinse rag, observation: you successfully rinse rag. Sink is CLEAN, faucet is ON and CLEAN, rag is CLEAN, faucet ON sink, faucet ON kitchen\_counter. You have rag.)]}\par
\vspace{0.3em}

\end{memorybox}

\caption{Example of the folded subgoal-level memory used by HIPIF on VirtualHome.}
\label{fig:folded_memory_virtualhome}
\end{figure}

\begin{figure}[htbp]
\centering
\begin{memorybox}{Folded History Example on ScienceWorld}
\scriptsize

\textbf{Your task is to:} measure the temperature of wood, which is located around the living room. First, focus on the thermometer. Next, focus on the wood. If the wood temperature is above 200.0 degrees celsius, place it in the green box. If the wood temperature is below 200.0 degrees celsius, place it in the blue box. The boxes are located around the living room.\par
\textbf{Your current subgoal is:} place the wood in the blue box.\par
\vspace{0.4em}

\strong{EXECUTION HISTORY:}\par
Subgoals are essential milestones on the path to the final task. Below is the execution history, each entry enclosed in \texttt{[]} contains a subgoal followed by all corresponding actions and observations:\par
\vspace{0.3em}

{[(\textbf{current\_subgoal}: place the wood in the blue box),}\par
{(pre\_action: move wood to blue box, observation: You move the wood to the blue box.)]}\par
\vspace{0.3em}

{[(\textbf{origin\_observation}: This room is called the foundry. In it, you see: the agent, a substance called air, a blast furnace, which is turned off. The blast furnace door is closed. A sink, which is turned off. In the sink is: nothing. A table. On the table is: nothing. You also see: A door to the outside that is open.)]}\par
\vspace{0.3em}

{[(\textbf{history\_subgoal 1}: find a thermometer),}\par
{(pre\_action: teleport to kitchen, observation: You teleport to the kitchen. This room is called the kitchen. In it, you see: the agent, a substance called air, a chair, a counter, a bowl containing a red apple, a banana, an orange, and a potato, a drawer, a cupboard, a finger painting, a freezer, a fridge, a glass jar containing sodium chloride, a lighter, an oven, a sink, soap, a stopwatch, a stove, a table, a glass cup, and a thermometer currently reading a temperature of 10 degrees celsius.)]}\par
\vspace{0.3em}

{[(\textbf{history\_subgoal 2}: pick up a thermometer),}\par
{(pre\_action: pick up thermometer, observation: You move the thermometer to the inventory.)]}\par
\vspace{0.3em}

{[(\textbf{history\_subgoal 3}: find the wood),}\par
{(pre\_action: focus on wood, observation: You foucus on wood.)]}\par
\vspace{0.3em}

{[(\textbf{history\_subgoal 4}: check the temperature of the wood),}\par
{(pre\_action: use thermometer on wood, observation: the thermometer measures a temperature of 4 degrees celsius)]}\par
\vspace{0.3em}

\end{memorybox}
\caption{Example of the folded subgoal-level memory used by HIPIF on ScienceWorld.}
\label{fig:folded_memory_scienceworld}
\end{figure}

\section{Validation Success Rate of 3B and 7B Models}
\label{app:val_curve_main}

\begin{figure*}[htbp]
\centering
\begin{subfigure}[b]{0.32\textwidth}
    \centering
    \includegraphics[width=\textwidth]{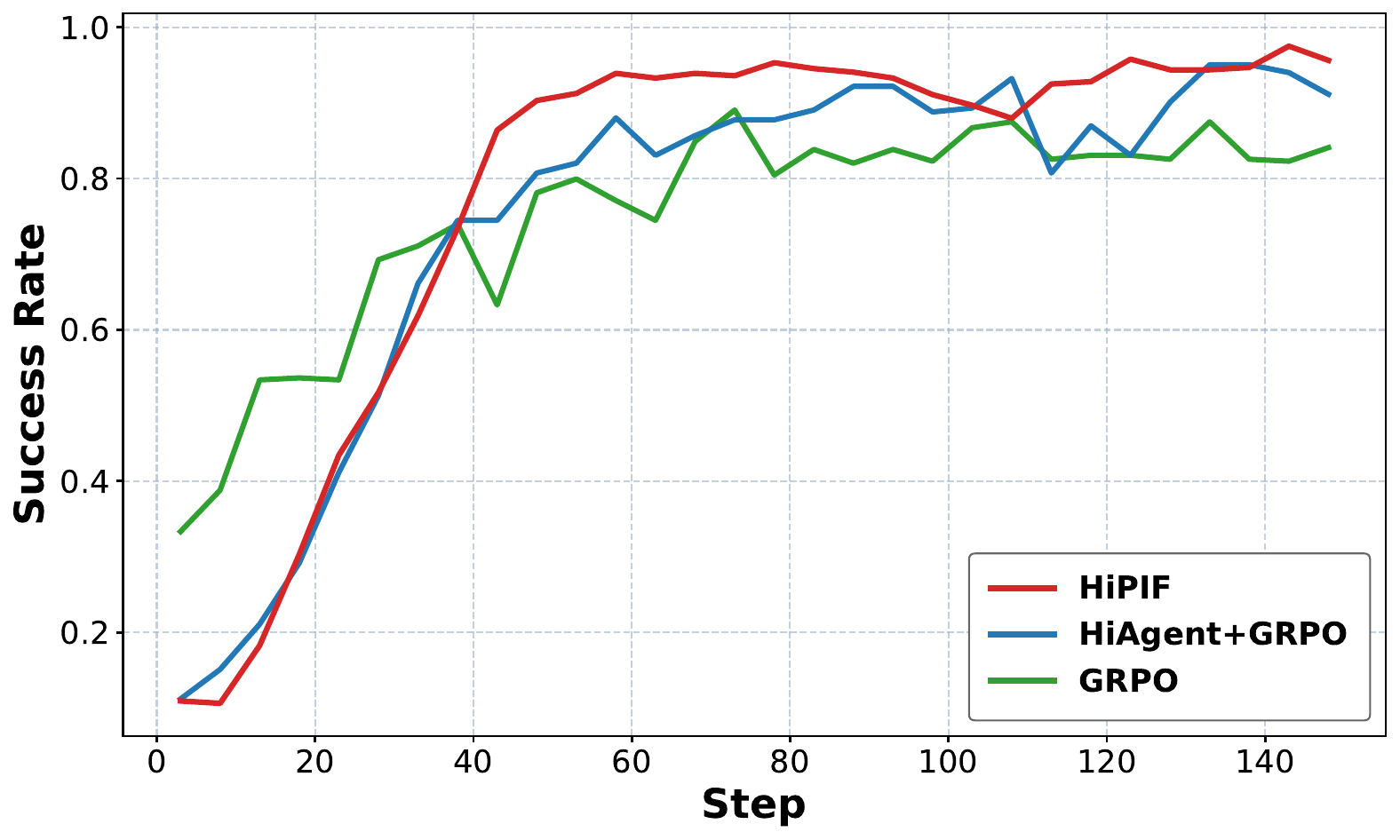}
    \caption{ALFWorld}
\end{subfigure}
\hfill
\begin{subfigure}[b]{0.32\textwidth}
    \centering
    \includegraphics[width=\textwidth]{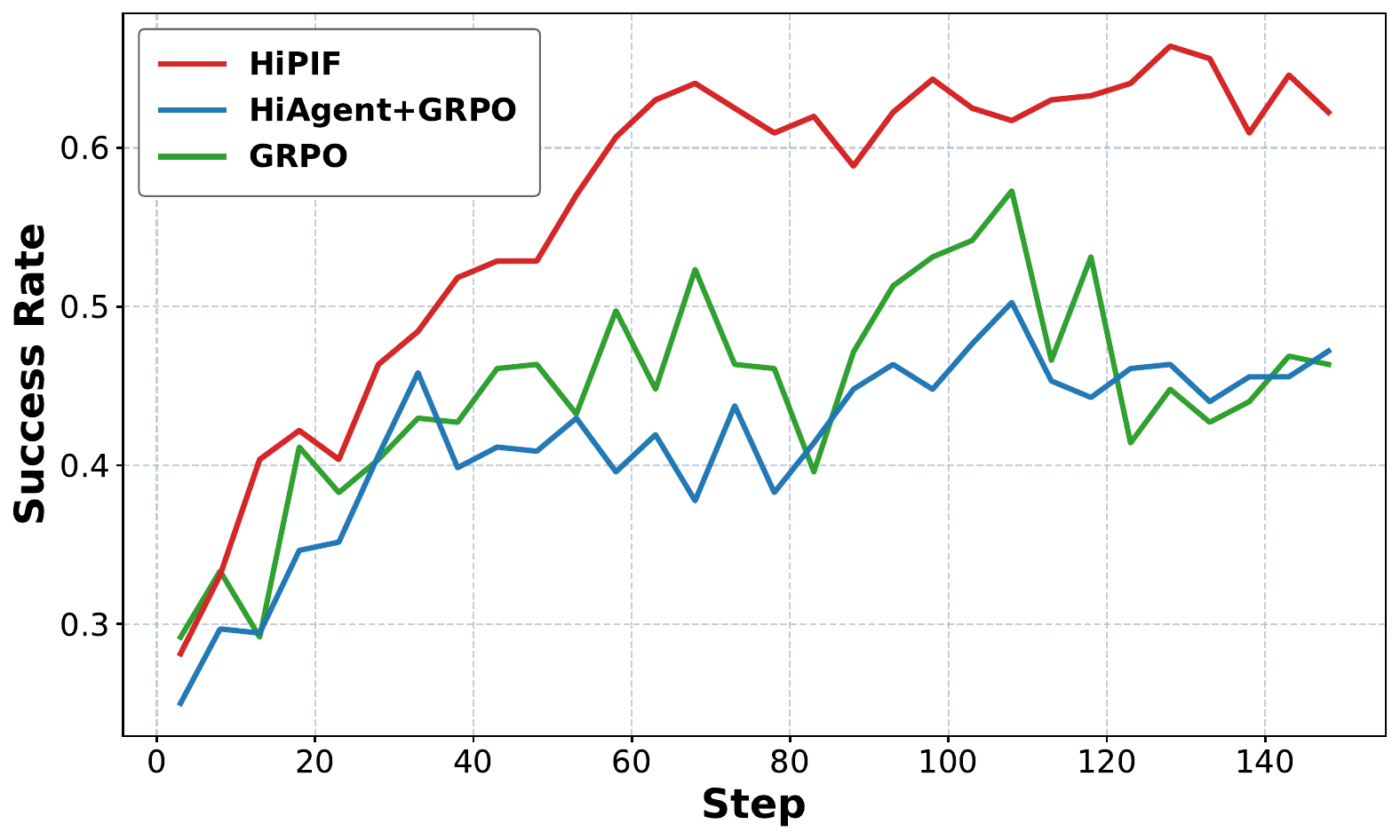}
    \caption{VirtualHome}
\end{subfigure}
\hfill
\begin{subfigure}[b]{0.32\textwidth}
    \centering
    \includegraphics[width=\textwidth]{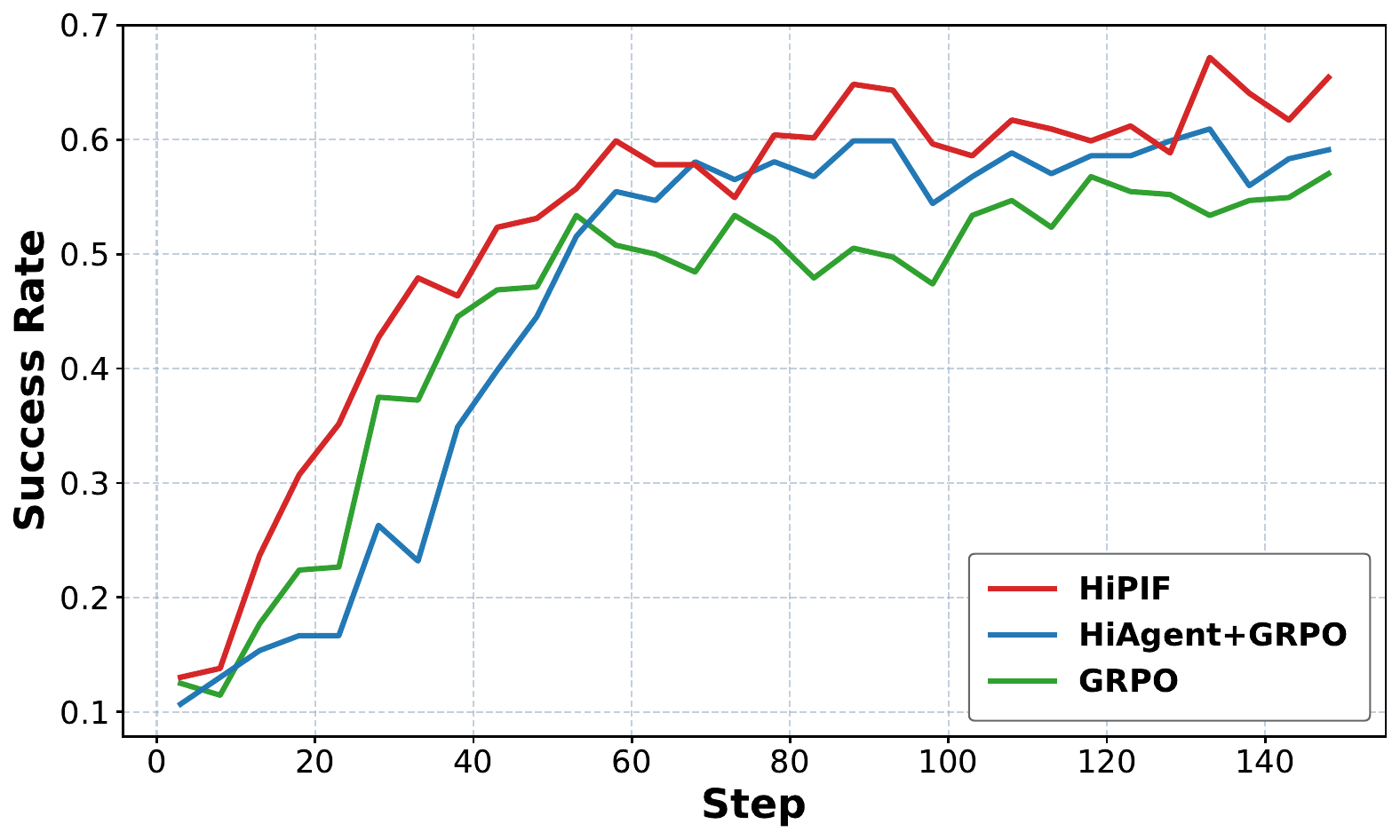}
    \caption{ScienceWorld}
\end{subfigure}
\caption{Validation success-rate curves of 7B models on ALFWorld, VirtualHome, and ScienceWorld.}
\label{fig:val_curve_7b}
\end{figure*}

To further examine the training dynamics of different methods, we present the validation success-rate curves of 3B and 7B models in Figure~\ref{fig:val_curve_3b} and Figure~\ref{fig:val_curve_7b}, respectively. 
Across all three benchmarks, HIPIF consistently achieves the strongest validation performance, while HiAgent+GRPO remains consistently above the GRPO baseline. 
This trend is stable for both model scales, indicating that the advantage of HIPIF is not limited to a specific parameter size. 

More importantly, the curves show that the gains of HIPIF are not only reflected in the final validation accuracy, but also in the overall training trajectory. 
Compared with GRPO, subgoal-based training can be less stable at the early stage, indicating that explicit subgoals introduce additional optimization difficulties. 
The later improvement of HIPIF shows that hierarchical reflection and fine-grained process rewards effectively stabilize subgoal-based RL and lead to better optimization behavior.

\section{Complementarity with GiGPO}
\label{app:val_curve_gigpo}

\begin{figure*}[htbp]
\centering
\begin{subfigure}[b]{0.32\textwidth}
    \centering
    \includegraphics[width=\textwidth]{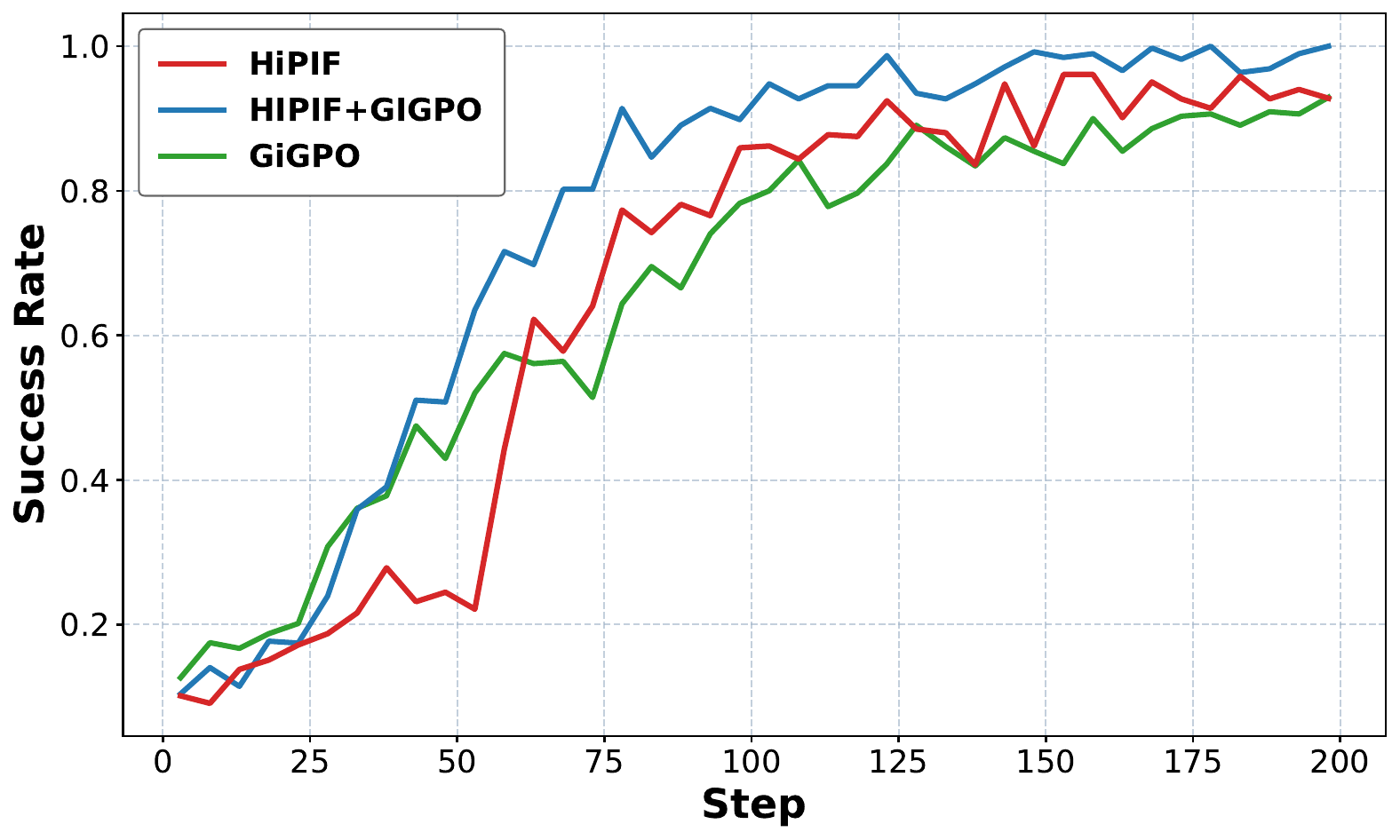}
    \caption{ALFWorld}
\end{subfigure}
\hfill
\begin{subfigure}[b]{0.32\textwidth}
    \centering
    \includegraphics[width=\textwidth]{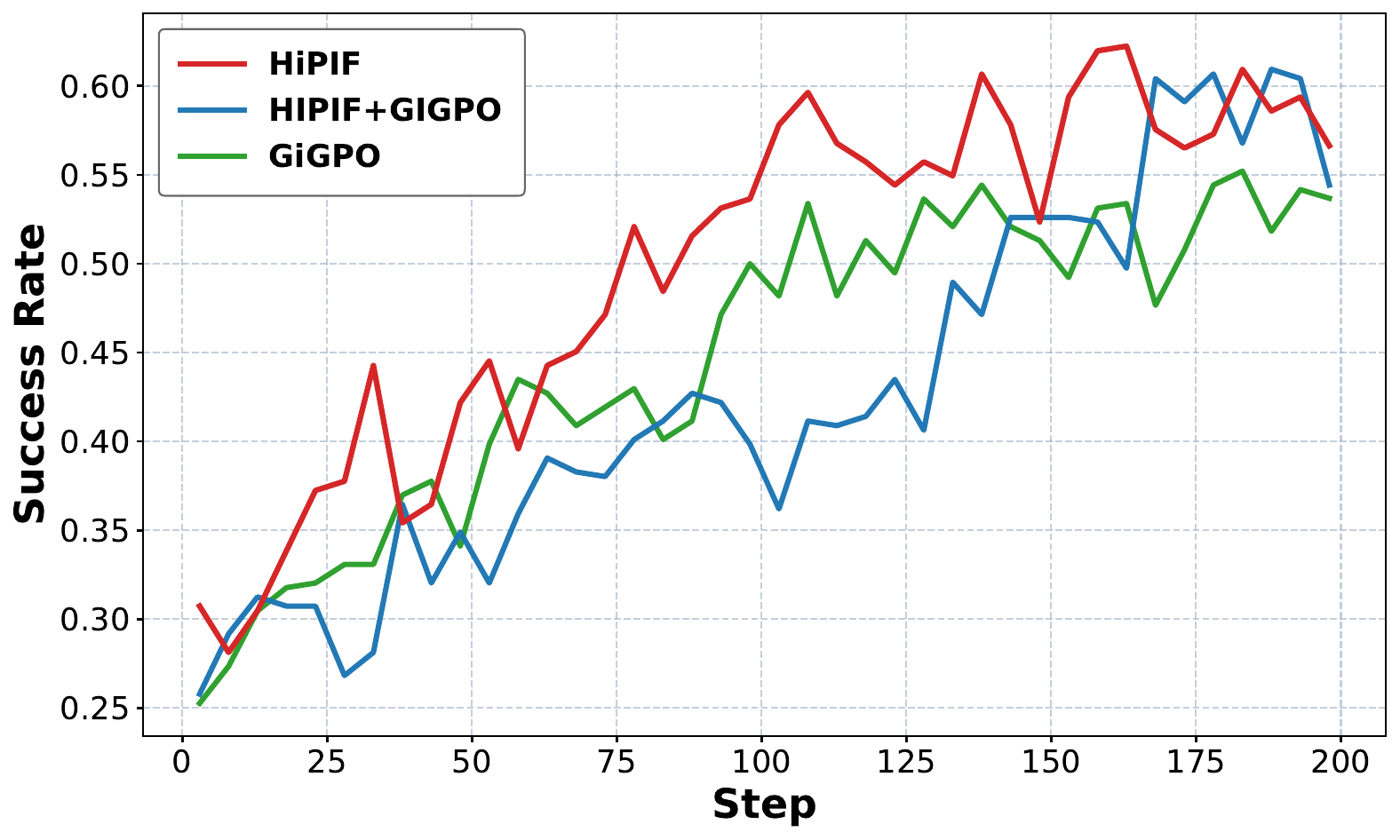}
    \caption{VirtualHome}
\end{subfigure}
\hfill
\begin{subfigure}[b]{0.32\textwidth}
    \centering
    \includegraphics[width=\textwidth]{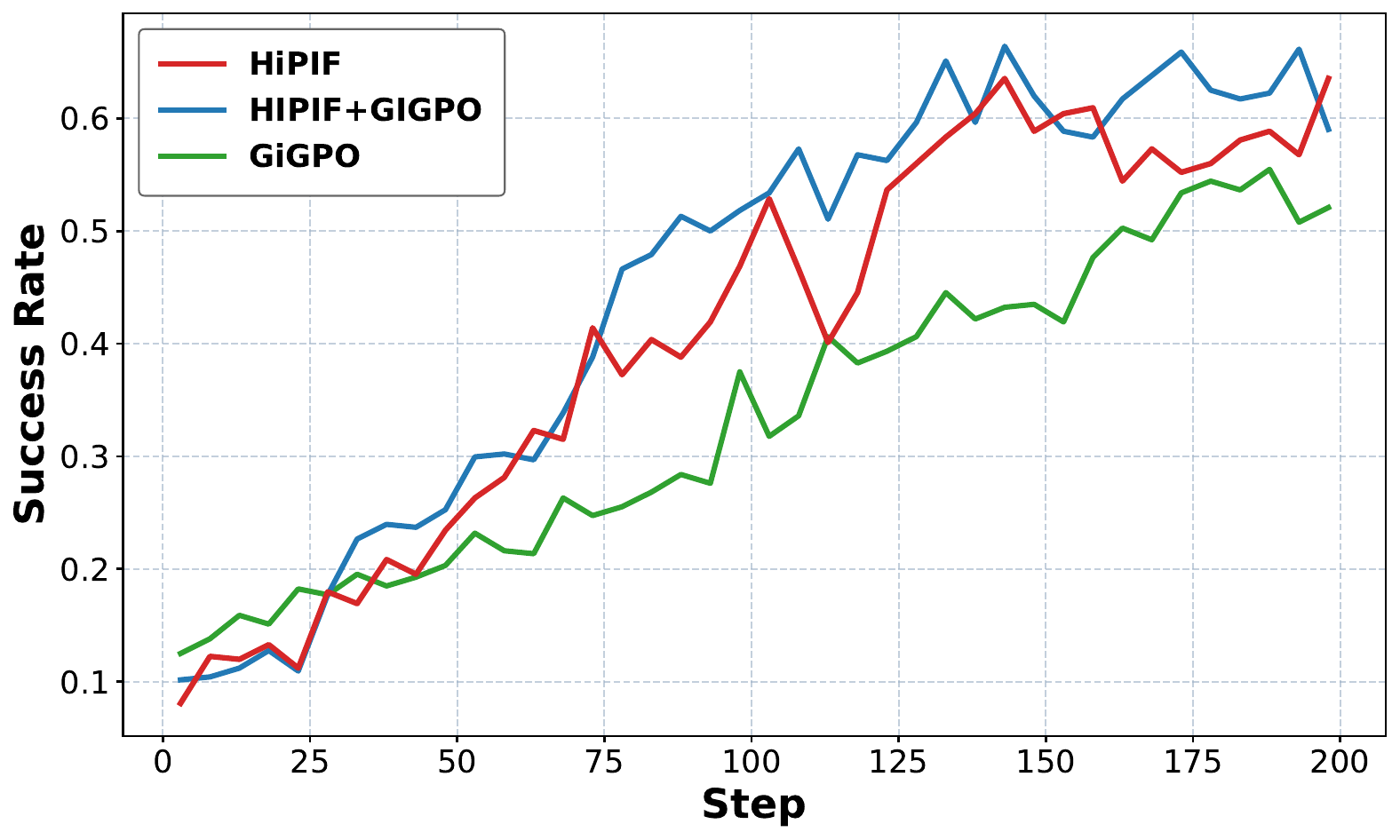}
    \caption{ScienceWorld}
\end{subfigure}
\caption{Validation success-rate curves comparing GiGPO, HIPIF, and HIPIF+GiGPO on the 3B setting.}
\label{fig:gigpo_curve_row}
\end{figure*}

We further study the relationship between HIPIF and GiGPO through validation success-rate curves on the 3B setting, as shown in Figure~\ref{fig:gigpo_curve_row}. 
Across all three benchmarks, HIPIF+GiGPO performs best, HIPIF ranks second, and GiGPO alone performs worst. 

More specifically, GiGPO primarily improves optimization through enhanced group-relative credit assignment, whereas HIPIF restructures the decision process with explicit hierarchical planning  and context folding, and further stabilizes subgoal-based training through hierarchical reflection and subgoal-oriented process rewards.
Therefore, combining HIPIF with GiGPO yields the strongest validation performance, indicating that the two methods are largely complementary. At the same time, HIPIF alone still outperforms GiGPO across all three benchmarks, showing that the structural improvements introduced by HIPIF provide stronger gains than using GiGPO alone.

\begin{table}[htbp]
\centering
\caption{Summary of the orthogonality analysis between HIPIF and GiGPO, based on the validation curves in Figure~\ref{fig:gigpo_curve_row}.}
\label{tab:gigpo_orthogonality}
\vspace{1mm}
\small
\setlength{\tabcolsep}{8pt}
\renewcommand{\arraystretch}{1.05}
\begin{tabular}{lccc}
\toprule
Methods & ALFWorld & VirtualHome & ScienceWorld \\
\midrule
HIPIF+GiGPO & \textbf{98.4} & \textbf{63.6} & \textbf{67.4} \\
HIPIF & 96.1 & 63.3 & 64.8 \\
GiGPO & 93.8 & 60.9 & 58.1 \\
\bottomrule
\end{tabular}
\end{table}

\section{Limitations.}
\label{app:limitation}
HIPIF is primarily evaluated in simulated long-horizon interaction benchmarks with structured observations and action spaces. While these environments provide controlled and reproducible testbeds, extending the framework to more open-ended real-world settings may require additional perception and action-grounding components. Moreover, HIPIF uses structured outputs for subgoal proposal, reflection, and execution, which improves interpretability but may require sufficiently capable instruction-following models. Finally, although we evaluate HIPIF on multiple benchmarks and two backbone scales, our experiments do not exhaustively cover all model families, parameter scales, or training budgets. Further evaluation on more diverse backbone models would provide a more comprehensive understanding of its scaling behavior.

\section{Case Studies.}
\label{app:case_studies_addition}
To provide a more intuitive understanding of HIPIF, we present additional case studies on VirtualHome and ScienceWorld in Tables~\ref{tab:case_study_virtualhome} and~\ref{tab:case_study_scienceworld}. These examples compare HIPIF with two ablated variants, \textbf{w/o Subgoal} and \textbf{w/o Reflection}, at representative key decision points in long-horizon tasks. Overall, the case studies show that subgoal decomposition and context folding help HIPIF preserve completed progress while avoiding distraction from long execution histories, while hierarchical reflection enables reliable assessment of subgoal completion, action failure, and subgoal transition. 

\begin{table}[htbp]
\centering
\caption{Case study on a long-horizon task in VirtualHome.}
\label{tab:case_study_virtualhome}
\vspace{1mm}
\scriptsize
\setlength{\tabcolsep}{1.8pt}
\renewcommand{\arraystretch}{1.02}
\resizebox{\textwidth}{!}{%
\begin{tabular}{L{1.0cm}L{4.3cm}L{4.3cm}L{4.3cm}}
\toprule
\textbf{Method} 
& \multicolumn{1}{c}{\textbf{HIPIF}} 
& \multicolumn{1}{c}{\textbf{w/o Subgoal}} 
& \multicolumn{1}{c}{\textbf{w/o Reflection}} \\
\midrule
\textbf{Task}
&
\multicolumn{3}{L{13.5cm}}{
 Wipe down counter. Instructions: Walk to dining room. Grab the rag, rinse it in the sink using the faucet, and then
wipe the kitchen counter.
}
\\
\midrule
\textbf{History}
&
\textbf{Subgoal 1}: walk to dining room. 
Res 1: you successfully walk to dining\_room, you see a rag, a faucet.
\newline
\textbf{Subgoal 2}: find and grab the rag.
Res 2: you successfully grab rag.
\newline
\textbf{Subgoal 3}: find the sink.
Res 3: you successfully walk to sink.
\newline
\textbf{Current\_Subgoal}: find the faucet.
Action: go to bathroom.
Obs: you successfully walk to bathroom, you see toilet, tooth\_paste, toothbrush. 
\par
\textcolor{omitgray}{\textbf{[... omitted following steps ...]}}
\par
&
\textbf{Action 1}: walk to dining room. 
Obs 1: you successfully walk to dining\_room, you see a rag.
\newline
\textbf{Action 2}: walk to the rag.
Obs 2: you successfully walk to the rag.
\newline
\textbf{Action 3}: grab the rag.
Obs 3: you successfully grab rag.
\par
\textcolor{omitgray}{\textbf{[... omitted intermediate steps ...]}}
\par
\textbf{Action 28}: walk to the rag.
Obs 28: Nothing happens.
\newline
\textbf{Action 29}: grab the rag.
Obs 29: Nothing happens.
\par
\textcolor{omitgray}{\textbf{[... omitted following steps ...]}}
\par
&
\textbf{Subgoal 1}: walk to dining room. 
Res 1: you successfully walk to dining\_room, you see a rag, a faucet.
\newline
\textbf{Subgoal 2}: find and grab the rag.
Res 2: you successfully grab rag.
\newline
\textbf{Subgoal 3}: find the sink.
Res 3: you successfully walk to sink.
\newline
\textbf{Current\_Subgoal}: find the faucet.
Action: go to bathroom.
Obs: you successfully walk to bathroom, you see toilet, tooth\_paste, toothbrush. 
\par
\textcolor{omitgray}{\textbf{[... omitted following steps ...]}}
\par
\\

\midrule

\textbf{Key step}
&
\textcolor{successgreen}{\textbf{Reflection:} Part1: The observation shows that on the bathroom, there is no faucet. The current subgoal is not completed. Part2: The next action should be to go to the dining\_room to check if there is a faucet.
\newline
\textbf{Action:} go to dining\_room.}
\newline
\textbf{Observation:} you successfully walk to dining\_room, you see a faucet.
&
\textcolor{failred}{\textbf{Thinking:} Based on the previous observations, I should use the rag to wipe. So I should first try to grab the rag. (the agent has already contained the rag) 
\newline
\textbf{Action:} grab the rag.}
\newline
\textbf{Observation:} Nothing happens.
&
\textcolor{failred}{\textbf{Reflection:} I have successfully see a faucet in bathroom. Now, I need to walk to the faucet and the rinse the rag using it.  (the agent now in bathroom, but faucet is in dining room)
\newline
\textbf{Subgoal:} Walk to faucet.
\newline
\textbf{Action:} walk to faucet.}
\newline
\textbf{Observation:} Nothing happens.
\\
\midrule

\textbf{Outcome}
&
\textcolor{successgreen}{\textbf{Success.}}
&
\textcolor{failred}{\textbf{Failure.}}
&
\textcolor{failred}{\textbf{Failure.}}
\\

\bottomrule
\end{tabular}
}
\end{table}

\begin{table}[htbp]
\centering
\caption{Case study on a long-horizon task in ScienceWorld.}
\label{tab:case_study_scienceworld}
\vspace{1mm}
\scriptsize
\setlength{\tabcolsep}{1.8pt}
\renewcommand{\arraystretch}{1.02}
\resizebox{\textwidth}{!}{%
\begin{tabular}{L{1.0cm}L{4.3cm}L{4.3cm}L{4.3cm}}
\toprule
\textbf{Method} 
& \multicolumn{1}{c}{\textbf{HIPIF}} 
& \multicolumn{1}{c}{\textbf{w/o Subgoal}} 
& \multicolumn{1}{c}{\textbf{w/o Reflection}} \\
\midrule
\textbf{Task}
&
\multicolumn{3}{L{13.5cm}}{
Measure the temperature of wood, which is located around the living room. First, focus on the thermometer. Next, focus on the wood. If the wood temperature is above 200.0 degrees celsius, place it in the green box. If the wood temperature is below 200.0 degrees celsius, place it in the blue box. The boxes are located around the living room.
}
\\
\midrule
\textbf{History}
&
\textbf{Subgoal 1}: find a thermometer. 
Res 1: You move the thermometer to the inventory.
\newline
\textbf{Subgoal 2}: find the wood.
Res 2: You foucus on wood.
\newline
\textbf{Subgoal 3}: check the temperature of the wood.
Res 3: the thermometer measures a temperature of 4 degrees celsius.
\newline
\textbf{Current\_Subgoal}: place the wood in the blue box.
Action: move wood to blue box.
Obs: Nothing happens. 
\par
\textcolor{omitgray}{\textbf{[... omitted following steps ...]}}
\par
&
\textbf{Action 1}: go to foundry. 
Obs 1: This room is called the foundry. In it, you see: A door to the outside that is open.
\newline
\textbf{Action 2}: go to outside.
Obs 2: This room is called the foundry. In it, you see: A door to the kitchen that is open.
\newline
\textbf{Action 3}: go to kitchen.
Obs 3: This room is called the kitchen. In it, you see: a drawer, a cupboard, a finger
painting, a freezer, a fridge, and a thermometer.
\par
\textcolor{omitgray}{\textbf{[... omitted intermediate steps ...]}}
\par
\textbf{Action 28}: go to outside.
Obs 28: This room is called the foundry. In it, you see: A door to the kitchen that is open.
\newline
\textbf{Action 29}: go to outside.
Obs 29: Nothing happens.
\par
\textcolor{omitgray}{\textbf{[... omitted following steps ...]}}
\par
&
\textbf{Subgoal 1}: find a thermometer. 
Res 1: You move the thermometer to the inventory.
\newline
\textbf{Subgoal 2}: find the wood.
Res 2: You foucus on wood.
\newline
\textbf{Subgoal 3}: check the temperature of the wood.
Res 3: the thermometer measures a temperature of 4 degrees celsius.
\newline
\textbf{Current\_Subgoal}: place the wood in the blue box.
Action: move wood to blue box.
Obs: Nothing happens. 
\par
\textcolor{omitgray}{\textbf{[... omitted following steps ...]}}
\par
\\

\midrule

\textbf{Key step}
&
\textcolor{successgreen}{\textbf{Reflection:} Part1: The observation shows the current subgoal is not completed. Part2: The last action failed because I was not close to the blue box. The next action should be to go to the kitchen to find the blue box.
\newline
\textbf{Action:} go to kitchen.}
\newline
\textbf{Observation:} you successfully walk to kitchen, you see a green box and a blue box.
&
\textcolor{failred}{\textbf{Thinking:} Based on the observations, I should first find the thermometer. The thermometer may be in outside.
\newline
(the agent has lost in long contexts) 
\newline
\textbf{Action:} go to outside.}
\newline
\textbf{Observation:} Nothing happens.
&
\textcolor{failred}{\textbf{Reflection:} The  temperature of the wood is 4 degrees celsius.I should move the wood to the blue box. 
\newline
(the agent falls into a loop.)
\newline
\textbf{Subgoal:} Walk to faucet.
\newline
\textbf{Action:} move wood to blue box.}
\newline
\textbf{Observation:} Nothing happens.
\\
\midrule

\textbf{Outcome}
&
\textcolor{successgreen}{\textbf{Success.}}
&
\textcolor{failred}{\textbf{Failure.}}
&
\textcolor{failred}{\textbf{Failure.}}
\\

\bottomrule
\end{tabular}
}
\end{table}


\clearpage

\end{document}